\documentclass[twocolumn]{article}%
\usepackage{amsfonts}
\usepackage{amsmath}
\usepackage{amssymb}
\usepackage{graphicx}%
\begin{document}

\title{ Contrast Trees and Distribution Boosting}
\author{Jerome H. Friedman$\thanks{Department of Statistics, Stanford University,
Stanford, 94305, USA.(jhf@stanford.edu)}$}
\date{}
\maketitle

\begin{abstract}
A new method for decision tree induction is presented. Given a set of
predictor variables $\mathbf{x}=(x_{1},x_{2},\cdot\cdot\cdot,x_{p})$ and
\emph{two} outcome variables $y$ and $z$ associated with each $\mathbf{x}$,
the goal is to identify those values of $\mathbf{x}$ for which the respective
distributions of \ $y\,|\,\mathbf{x}$ and \ $z\,|\,\mathbf{x}$, or selected
properties of those distributions such as means or quantiles, are most
different. Contrast trees provide a lack-of-fit measure for statistical models
of such statistics, or for the complete conditional distribution
$p_{y}(y\,|\,\mathbf{x})$, as a function of $\mathbf{x}$. They are easily
interpreted and can be used as diagnostic tools to reveal and then understand
the inaccuracies of models produced by any learning method. \ A corresponding
contrast boosting strategy is described for remedying any uncovered errors
thereby producing potentially more accurate predictions. This leads to a
distribution boosting strategy for directly estimating the full\ conditional
distribution of $y$ at each $\mathbf{x}$\ under no assumptions concerning its
shape, form or parametric representation.

\ 


\ \ 

\end{abstract}

\section{Introduction}

In statistical (machine) learning one has a system under study with associated
attributes or variables. The goal is to estimate the unknown value of one of
the variables $y$, given the known joint values of other (predictor) variables
$\mathbf{x}=$ $(x_{1}\cdot\cdot\cdot,x_{p})$ associated with the system. It is
seldom the case that a particular set of $\mathbf{x}$--values gives rise to a
unique value for $y$. There are quantities other than those in $\mathbf{x}$
that influence $y$ whose values are neither controlled nor observed.
Specifying a particular set of joint values for $\mathbf{x}$ results in a
probability distribution of possible $y$-values, $p_{y}(y\,|\,\mathbf{x})$,
induced by the varying values of the uncontrolled quantities. Given a sample
$\{y_{i},\mathbf{x}_{i}\}_{i=1}^{N}$ of previous solved cases, the goal is to
estimate the distribution $p_{y}(y\,|\,\mathbf{x})$, or some of its
properties, as a function of the predictor variables $\mathbf{x}$. These can
then be used to predict likely values of $y$ realized at each $\mathbf{x}$.

Usually only a single property of $p_{y}(y\,|\,\mathbf{x})$ is used for
prediction, namely a measure of its central tendency such as the mean or
median. This provides no information concerning individual accuracy of each
prediction. Only average accuracy over a set of predictions (cross-validation)
can be determined. In order to estimate individual prediction accuracy at each
$\mathbf{x}$ one needs additional properties of $p_{y}(y\,|\,\mathbf{x})$ such
as various quantiles, or the distribution itself. These can be estimated as
functions of $\mathbf{x}$ using maximum likelihood or minimum risk techniques.
Such methods however do not provide a measure of accuracy (goodness-of-fit)
for their respective estimates as functions of $\mathbf{x}$. There is no way
to know how well the results actually \ characterize the distribution of $y$
at each $\mathbf{x}$.

Contrast trees can be used to assess\ lack-of-fit of any estimate of
$p_{y}(y\,|\,\mathbf{x})$, or its properties (mean, quantiles), as a function
of $\mathbf{x}$. In cases where the fit is found to be lacking, contrast
\emph{boosting} applied to the output can often improve accuracy. A special
case of contrast boosting, \emph{distribution }boosting, can be used to
estimate the full conditional distribution $p_{y}(y\,|\,\mathbf{x}\mathbb{)}$
under no assumptions. Contrast trees can also be used to uncover concept drift
and reveal discrepancies in the predictions of different learning algorithms.

\section{Contrast trees}

A classification or regression tree partitions the space of $\mathbf{x}$ -
values into easily interpretable regions defined by simple conjunctive rules.
The goal is to produce regions in $\mathbf{x}$ - space such that the variation
of $y$ values within each is made small. A \emph{contrast} tree also
partitions the $\mathbf{x}$ - space into similarly defined regions, but with a
different purpose. There are \emph{two} outcome variables $y$ and $z$
associated with each $\mathbf{x}$. The goal is to find regions in $\mathbf{x}$
- space where the values of the two variables are most different.

In some applications of contrast trees the outcomes $y$ and $z$ can be
different functions of $\mathbf{x}$, $y=f(\mathbf{x})$ and $z=g(\mathbf{x}%
)$,\ such as predictions produced by two different learning algorithms. The
goal of the contrast tree is then to identify regions in $\mathbf{x}$ - space
where the two predictions most differ. In other cases the outcome $y$ may be
observations of a random variable assumed to be drawn from some distribution
at $\mathbf{x}$, $y\sim p_{y}(y\,|\,\mathbf{x})$. The quantity $z$ might be an
estimate for some property of that distribution such as its estimated mean
$\hat{E}(y\,|\,\mathbf{x})$ or $p$-th quantile $\hat{Q}_{p}(y\,\,|\mathbf{\,x}%
)$ as a function of $\mathbf{x}$. One would like to identify $\mathbf{x}$ -
values where the estimates $z$ appear to be the least accurate. Alternatively
$z$ itself could be a random variable independent of $y$ (given $\mathbf{x}$)
with distribution $p_{z}(z\,|\,\mathbf{x})$ and interest is in identifying
regions of $\mathbf{x}$ - space where the two distributions $p_{y}%
(y\,|\,\mathbf{x})$ and $p_{z}(z\,|\,\mathbf{x})$ most differ.

In these applications contrast trees can be used as diagnostics to ascertain
the lack-of-fit of statistical models to data or to other models. As with
other tree based methods the uncovered regions are defined by conjunctive
rules based on \ simple logical statements concerning the variable values.
Thus it is straightforward to understand the joint predictor variable values
at which discrepancies have been identified. Such information may temper
confidence in some predictions or suggest ways to improve accuracy.

In prediction problems $z$ is taken to be an estimate of some property of the
distribution $p_{y}(y\,|\,\mathbf{x})$, or of the distribution itself. One way
to improve accuracy is to modify the predicted values $z$ in a way that
reduces their discrepancy with the actual values as represented by the data.
Contrast trees attempt to identify regions of $\mathbf{x}$ - space with the
largest discrepancies. The $z$ - values within in each such region can then be
modified to reduce discrepancy with $y$. This produces new values of $z$ which
can then be contrasted with $y$ using another contrast tree. This process can
then be applied to the regions of the new tree thereby producing further
modified $z$ - values. This \textquotedblleft boosting\textquotedblright%
\ strategy of successively building contrast trees on the output of previously
induced trees can be continued until the average discrepancy stops improving.

\section{Building contrast trees\label{s2}}

The data consists of $N$ observations $\{\mathbf{x}_{i},y_{i},z_{i}%
\}_{i=1}^{N\ }$ each with a joint set of predictor variable values
$\mathbf{x}_{i}$ and two outcome variables $y_{i}$ and $z_{i}$. Contrast trees
are constructed from this data in an iterative manner. At the $M$th iteration
the tree partitions the space of $\mathbf{x}$ - values into $M$ disjoint
regions $\{R_{m}\}_{m=1}^{M}$ each containing a subset of the data
$\{\mathbf{x}_{i},y_{i},z_{i}\}_{\mathbf{x}_{i}\in R_{m}}$. At the first
iteration there is a single region containing the entire data set. Associated
with any data subset is a discrepancy measure between the $y$ and $z$ values
of the subset
\begin{equation}
d_{m}=D(\{y_{i}\}_{\mathbf{x}_{i}\in R_{m}},\{z_{i}\}_{\mathbf{x}_{i}\in
R_{m}})\text{.} \label{e1}%
\end{equation}
Choice of a particular discrepancy measure depends on the specific application
as discussed in Section \ref{s3}.

At the next ($M+1$)st iteration each of the regions $R_{m}$ defined at the
$M$th iteration \ ($1\leq m\leq M$) are provisionally partitioned (split) into
two regions $R_{m}^{(l)}$ and $R_{m}^{(r)}$ ($R(_{m}^{(l)}\cup$ $R_{m}%
^{(r)}=R_{m}$). Each of these \textquotedblleft daughter\textquotedblright%
\ regions contains its own data subset with associated discrepancy measure
$d_{m}^{(l)}$ and $d_{m}^{(r)}$ (\ref{e1}). The quality of the split is
defined as%
\begin{equation}
Q_{m}(l,r)=(f_{m}^{(l)}\cdot f_{m}^{(r)})\,\max(d_{m}^{(l)},d_{m}^{(r)})^{2}
\label{e2}%
\end{equation}
where $f_{m}^{(l)}$ and $f_{m}^{(r)}$ are the fraction of observations in the
\textquotedblleft parent\textquotedblright\ region $R_{m}$ associated with
each of the two daughters.

The types of splits considered here are the same as in ordinary classification
and regression trees (Breiman \emph{et al} 1984). Each involves one of the
predictor variables $x_{j}$. For numeric variables splits are specified by a
particular value of that variable (split point) $s$. The corresponding
daughter regions are defined by%
\begin{align}
\mathbf{x}  &  \in R_{m}\,\&\,x_{j}\leq s\Longrightarrow\mathbf{x}\in
R_{m}^{(l)}\label{e3}\\
\mathbf{x}  &  \in R_{m}\,\&\,x_{j}>s\Longrightarrow\mathbf{x}\in R_{m}%
^{(r)}\text{.}\nonumber
\end{align}
For categorical variables (factors) the respective levels are ordered by
discrepancy (\ref{e1}). The discrepancy at each respective level of the factor
for the observations in the $m$th region is computed. Splits are then
considered in this order.

Split quality (\ref{e2}) is evaluated jointly for all current regions $R_{m}$,
all predictor variables, and all possible splits of each variable. The region
with largest estimated split improvement%
\begin{equation}
I_{m}=\max(d_{m}^{(l)},d_{m}^{(r)})-d_{m} \label{e4}%
\end{equation}
is then chosen to create two new regions (\ref{e3}). Here $d_{m}$ (\ref{e1})
is the discrepancy associated with the data in the parent region and
$d_{m}^{(l)},d_{m}^{(r)}$ are corresponding discrepancies of the data subsets
in the two daughters. These two new regions replace the corresponding parent
producing $M+1$ total regions. Splitting stops when no estimated improvement
(\ref{e4}) is greater than zero, the tree reaches a specified size or the
observation count within all regions is below a specified threshold.

\section{Discrepancy measures\label{s3}}

By defining different discrepancy measures contrast trees can be applied to a
variety of different problems. Even within a particular type of problem there
may be a number of different appropriate discrepancy measures that can be used.

When the two outcomes are simply functions of $\mathbf{x}$, $y=f(\mathbf{x})$
and $z=g(\mathbf{x})$, any quantity that reflects their difference in values
at the same $\mathbf{x}$ can be used to form a discrepancy measure such as%
\[
d_{m}=\frac{1}{N_{m}}\sum_{\mathbf{x}_{i}\in R_{m}}|\,y_{i}-z_{i}\,|\text{.}%
\]
If $y$ is a random variable and $z$ is an estimate for a property $S$ of its
conditional distribution at $\mathbf{x}$, $z_{i}=\hat{S}(p_{y}%
(y\,|\,\mathbf{x}_{i}))$, such as its mean $E(y\,|\,\mathbf{x}_{i})$ or $p$th
quantile $Q_{p}(y\,|\,\mathbf{x}_{i})$, a natural discrepancy measure is
\begin{equation}
d_{m}=\left\vert S(\{y_{i}\}_{\mathbf{x}_{i}\in R_{m}})-M(\{z_{i}%
\}_{\mathbf{x}_{i}\in R_{m}})\right\vert \text{.} \label{e5}%
\end{equation}
Here $S(\{y_{i}\}_{\mathbf{x}_{i}\in R_{m}})$ is the value of the
corresponding statistic computed for observations in the region $R_{m}$ and
$M(\{z_{i}\}_{\mathbf{x}_{i}\in R_{m}})$ is an appropriate measure of central
tendency for the corresponding estimates.

If both $y\sim p_{y}(y\,|\,\mathbf{x})$ and $z\sim p_{z}(z\,|\,\mathbf{x})$
are both independent random variables associated with each $\mathbf{x}$, a
discrepancy measure reflects the distance between their respective
distributions. There are many proposed empirical measures of distribution
distance. Every two--sample test has one. For the examples below a variant of
the Anderson--Darling (Anderson and Darling 1954) statistic is used. Let
$\{t_{i}\}=\{y_{i}\}\cup\,\{z_{i}\}$ represent the pooled $(y,z)$ sample in a
region $R_{m}$. Then discrepancy between the distributions of $y$ and $z$ is
taken to be%
\begin{equation}
d_{m}=\frac{1}{2N_{m}-1}\sum_{i=1}^{2N_{m}-1}\frac{\left\vert \hat{F}%
_{y}(t_{(i)})-\hat{F}_{z}(t_{(i)})\right\vert }{\sqrt{i\cdot(2N_{m}-i)}}
\label{e8}%
\end{equation}
where $t_{(i)}$ is the $i$th value of $t$ is sorted order, and $\hat{F}_{y}$
and $\hat{F}_{z}$ are the respective empirical cumulative distributions of $y$
and $z$. Note that this discrepancy measure (\ref{e8}) can be employed in the
presence of arbitrarily censored or truncated data simply by employing a
nonparametric method to estimate the respective CDF's such as Turnbul (1976) .

Discrepancy measures are often customized to particular individual
applications. In this sense they are similar to loss criteria in prediction
problems. However, in the context of contrast trees (and boosting) there is no
requirement that they be convex or even differentiable. Several such examples
are provided in the Appendix.

\section{Boosting contrast trees\label{s5}}

As indicated above, and illustrated in the examples presented below and in the
Appendix, contrast trees can be employed as diagnostics to examine the lack of
accuracy of predictive models. To the extent that inaccuracies are uncovered,
\emph{boosted} contrast trees can be used to attempt to mitigate
them,\ thereby producing more accurate predictions. Contrast boosting derives
successive modifications to an initially specified $z$, each reducing its
discrepancy with $y$ over the data. Prediction then involves starting with the
initial value of $z$ and then applying the modifications to produce the
resulting estimate.

\subsection{Estimation contrast boosting\label{s5.1}}

In this case $z$ is taken to be an estimate of some parameter of
$p_{y}(y\,|\,\mathbf{x})$. The $z$ - values within each region $R_{m}^{(1)}$
of a contrast tree can be modified $z\rightarrow z^{(1)}$ $=z+\delta_{m}%
^{(1)}\;(\mathbf{x}\in R_{m}^{(1)})$ so that the discrepancy (\ref{e1}) with
$y$ is zero in that region
\begin{equation}
D(\{y_{i}\}_{\mathbf{x}_{i}\in R_{m}^{(1)}},\{z_{i}^{(1)}\}_{\mathbf{x}_{i}\in
R_{m}^{(1)}})=0\text{.} \label{e13}%
\end{equation}
This in turn yields zero average discrepancy between $y$ and $z^{(1)}$ over
the regions defined by the terminal nodes of the corresponding contrast tree.
However, there may well be other partitions of the $\mathbf{x}$ - space
defining different regions $\{R_{m}^{(2)}\}_{1}^{M}$ for which this
discrepancy is not small. These may be uncovered by building a second tree to
contrast $y$ with $z^{(1)}$ producing updates%
\begin{equation}
z^{(2)}=z^{(1)}+\delta_{m}^{(2)}\;(\mathbf{x}\in R_{m}^{(2)}). \label{e13.5}%
\end{equation}
These in turn can be contrasted with $y$ to produce new regions $\{R_{m}%
^{(3)}\}_{1}^{M}$ and corresponding updates $\{\delta_{m}^{(3)}\}_{1}^{M}$.
Such iterations can be continued $K$ times until the updates become small. As
with gradient boosting (Friedman 2001) performance accuracy is often improved
by imposing a learning rate. At each step $k$\ the computed update $\delta
_{m}^{(k)}$ in each region $R_{m}^{(k)}$ is reduced by a factor $0<\alpha
\leq1$. That is, $\delta_{m}^{(k)}\leftarrow\alpha\,\delta_{m}^{(k)}$ in
(\ref{e13.5}).

Each tree $k$ in the boosted sequence $1\leq k\leq K$ partitions the
$\mathbf{x}$ - space into a set of regions $\{R_{m}^{(k)}\}$. Any point
$\mathbf{x}$ lies within one region $m_{k}(\mathbf{x})$ of each tree with
corresponding update $\delta_{m_{k}(\mathbf{x})}^{(k)}$. Starting with a
specified initial value $z_{0}(\mathbf{x})$ the estimate $\hat{z}(\mathbf{x})$
at $\mathbf{x}$ is then
\begin{equation}
\hat{z}(\mathbf{x})=z_{0}(\mathbf{x})+\sum_{k=1}^{K}\delta_{m_{k}(\mathbf{x}%
)}^{(k)}\text{.} \label{e14}%
\end{equation}

\subsection{Distribution contrast boosting\label{s5.2}}

Here $y$ and $z$ are both considered to be random variables independently
generated from respective distributions $p_{y}(y\,|\,\mathbf{x)}$ and
$p_{z}(z\,|\,\mathbf{x})$. The purpose of a contrast tree is to identify
regions of $\mathbf{x}$ - space where the two distributions most differ. The
goal of \emph{distribution} \emph{boosting} is to estimate a (different)
transformation of $z$ at each $\mathbf{x}$, $g_{\mathbf{x}}(z\,)$, such that
the distribution of the transformed variable is the same as that of $y$ at
$\mathbf{x}$. That is,%
\begin{equation}
p_{g_{\mathbf{x}}}(g_{\mathbf{x}}(z\,)\mathbf{\,|\,x})=p_{y}(y\,|\,\mathbf{x}%
)\text{.} \label{e15}%
\end{equation}
Thus, starting with $z$ values sampled from a known distribution
$p_{z}(z\,|\,\mathbf{x})$ at each $\mathbf{x}$, one can use the estimated
transformation $\hat{g}_{\mathbf{x}}(z\,)$ to obtain an estimate $\hat{p}%
_{y}(y\,|\,\mathbf{x})$ of the $y$ - distribution at that $\mathbf{x}$. Note
that the transformation $g_{\mathbf{x}}(z\,)$ is usually a different function
of $z$ at each different $\mathbf{x}$.

The $z$ - values within each region $R_{m}^{(1)}$ of a contrast tree can be
transformed $z^{(1)}$ $=g_{m}^{(1)}(z)\,\;(\mathbf{x}\in R_{m}^{(1)})$ so that
the discrepancy (\ref{e8}) with $y$ is zero in that region. The transformation
is given by%
\begin{equation}
g_{m}^{(1)}(z)=\hat{F}_{y}^{-1}\left(  \hat{F}_{z}(z)\right)  \label{e16}%
\end{equation}
where $\hat{F}_{y}(y)$ is the empirical cumulative distribution of $y$ for
$\mathbf{x}\in R_{m}^{(1)}$ and $\hat{F}_{z}(z)$ is the corresponding
distribution of $z$ for $\mathbf{x}\in R_{m}^{(1)}$. This transformation
function is represented by the quantile-quantile (QQ) plot of $y$ versus $z$
in the region.

As with estimation boosting, the distribution of the modified (transformed)
variable $z^{(1)}$ can then be contrasted with that of $y$ using another
contrast tree. This produces another region set $\{R_{m}^{(2)}\}_{1}^{M}$
where the distributions of $y$ and $z^{(1)}$ differ. This discrepancy
(\ref{e8}) can be removed by transforming $z^{(1)}$ to match the distribution
of $y$ in each new region $z^{(2)}$ $=g_{m}^{(2)}(z^{(1)})\,\;(\mathbf{x}\in
R_{m}^{(2)})$. These in turn can be contrasted with $y$ producing new regions
each with a corresponding transformation function. Such distribution boosting
iterations can be continued $K$ times until the discrepancy between the
distributions of $z^{(K)}$ and $y$ becomes small in each new region. As with
estimation, moderating the learning rate by shrinking each estimated
transformation function towards identity $g_{m}^{(k)}(z)\leftharpoonup
(1-\alpha)\,\,z+\alpha\,g_{m}^{(k)}(z)$ usually increases accuracy at the
expense of computation (more transformations).

Predicting $p_{y}(y\,|\,\mathbf{x})$ starts with a sample $\{z_{i}\}_{1}^{n}$
drawn from the specified distribution of $z$, $p_{z}(z\,|\,\mathbf{x})$, at
$\mathbf{x}$. This $\mathbf{x}$ lies within one of the regions $m_{k}%
(\mathbf{x})$ of each contrast tree\ $k$ with corresponding transformation
function $g_{m_{k}(\mathbf{x})}^{(k)}(\cdot)$. A given value of $z$ can be
transformed to a estimated value for $y$, $\hat{y}=\hat{g}_{\mathbf{x}}(z\,)$,
where%
\begin{equation}
\hat{g}_{\mathbf{x}}(z\,)=g_{m_{K}(\mathbf{x})}^{(K)}(g_{m_{K-1}(\mathbf{x}%
)}^{(K-1)}(g_{m_{K-2}(\mathbf{x})}^{(K-2)}\cdot\cdot\cdot g_{m_{1}%
(\mathbf{x})}^{(1)}(z)))\text{.} \label{e17}%
\end{equation}
That is, the transformed output of each successive tree is further transformed
by the next tree in the boosted sequence. A different transformation is chosen
at each step depending on the region of the corresponding tree containing
the\ particular joint values of the predictor variables $\mathbf{x}$. With $K$
trees each containing $M$ regions (terminal nodes) there are $M\,^{K}$
potentially different transformations $\hat{g}_{\mathbf{x}}(z\,)$ each
corresponding to different values of $\mathbf{x}$. To the extent the overall
transformation estimate $\hat{g}_{\mathbf{x}}(z\,)$ is accurate, the
distribution of the transformed sample $\{\hat{y}_{i}=\hat{g}_{\mathbf{x}%
}(z_{i}\,)\}_{1}^{n}$ can be regarded as being similar to that of $y$ at
$\mathbf{x}$, $p_{y}(y\,|\,\mathbf{x}\mathbb{)}$. Statistics computed from the
values of $\hat{y}$ estimating selected properties of its distribution, or the
distribution itself, can be regarded as estimates of the corresponding
quantities for $p_{y}(y\,|\,\mathbf{x}\mathbb{)}$.

\section{Diagnostics\label{s4}}

In this section we illustrate use of contrast trees as diagnostics for
uncovering and understanding the lack-of-fit of prediction models for
classification and conditional distribution estimation. Quantile regression
models are examined in the Appendix.

\subsection{Classification\label{s41}}

Contrast tree classification diagnostics are illustrated on the census income
data obtained from the Irvine Machine Learning repository (Kohvai 1996). This
data sample, taken from 1994 US census data, consists of observations from
48842 people divided into a training set of 32561 and an independent test set
of 16281. The outcome variable $y$ is binary and indicates whether\ or not a
person's income is greater than \$50000 per year. There are $14$ predictor
variables $\mathbf{x}$ consisting of various demographic and financial
properties associated with each person. Here we use contrast trees to diagnose
the predictions of gradient boosted regression trees (Friedman 2001).

The predictive model produced by the gradient boosting procedure applied to
the training data set produced an error rate of $13\%$ on the test data. This
quantity is the expected error as averaged over all test set predictions. It
may be of interest to discover certain $\mathbf{x}$ - values for which
expected error is much higher or lower. This can be ascertained by contrasting
the binary outcome variable $y$ with the model prediction $z$.

\begin{figure}
[ptb]
\begin{center}
\includegraphics[
height=3.5051in,
width=3.7014in
]%
{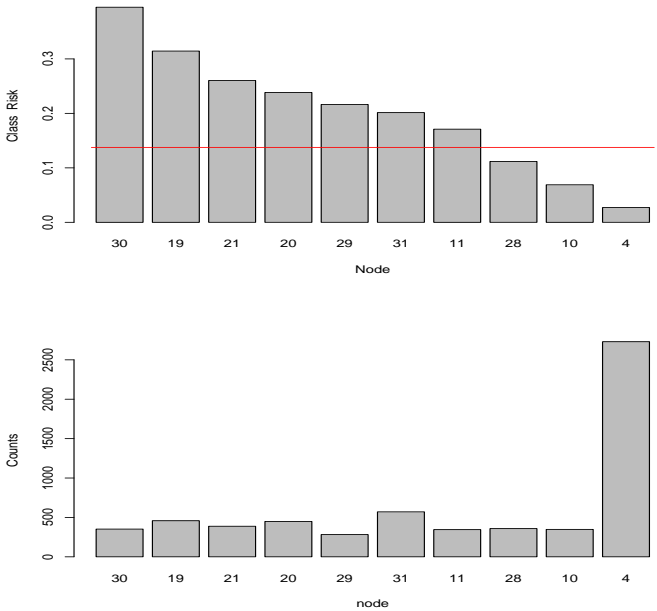}%
\caption{Error rate (upper) and observation count (lower) of classification
contrast tree regions on census income data. }%
\label{fig1}%
\end{center}
\end{figure}

A natural discrepancy measure for this application is error \ rate in each
region $R_{m}$%
\begin{equation}
d_{m}=\frac{1}{N_{m}}\sum_{i\in R_{m}}I(y_{i}\neq z_{i})\text{.} \label{e9}%
\end{equation}
The goal in applying contrast trees is to uncover regions in $\mathbf{x}$ -
space with exceptionally high values of (\ref{e9}). For this purpose the test
data set was randomly divided into two parts of 10000 \ and 6281 observations
respectively. The contrast tree procedure applied to the 10000 test data set
produced $10$ regions. Figure \ref{fig1} summarizes these regions using the
separate 6281 observation data set. The upper barplot shows the error rate of
the gradient boosting classifier in each region ordered from largest to
smallest. The lower barplot indicates the observation count in each respective
region. The number below each bar is simply the contrast tree node identifier
for that region. The horizontal (red) line indicates the $13\%$ average error rate.%

%TCIMACRO{\FRAME{ftbpFU}{3.7014in}{3.5051in}{0pt}{\Qcb{Error rate (upper) and
%observation count (lower) of classification contrast tree regions on census
%income data. }}{\Qlb{fig1}}{adultclass.eps}%
%{\special{ language "Scientific Word";  type "GRAPHIC";  display "USEDEF";
%valid_file "F";  width 3.7014in;  height 3.5051in;  depth 0pt;
%original-width 7.9952in;  original-height 10.5031in;  cropleft "0";
%croptop "1";  cropright "1";  cropbottom "0";
%filename 'adultclass.eps';file-properties "XNPEU";}} }%
%BeginExpansion
%EndExpansion

As Fig. \ref{fig1} indicates the contrast tree has uncovered many regions with
substantially higher error rates than the overall average and several others
with substantially lower error rates. The lowest error region covers $43\%$ of
the test set observations with an average error rate of $2.7\%$. The highest
error region covering $5.6\%$ of the data has an average error rate of $41\%$ .

Each of the regions represented in Fig. \ref{fig1} are easily described. For
example, the rule defining the lowest error region is

\begin{center}
\textbf{Node 4}

\thinspace

relationship $\in$ \{Own-child, Husband, Not-in-family, Other-relative\}

\&

education $\leq$ $12$

\thinspace
\end{center}

Predictions satisfying that rule suffer only a $2.7\%$ average error rate.
Predictions satisfying the rule defining the highest error region\newpage

\begin{center}
\textbf{Node 30}

\thinspace

relationship $\notin$ \{Own-child, Husband, Not-in-family, Other-relative\}

\&

occupation $\in$ \{ Exec-managerial, Transport-moving, Armed-Forces \}

\&

education $\leq$ $12$
\end{center}

have a $41\%$ average error rate. \ Thus confidence in salary predictions for
people in node 4 might be higher than for those in node 30.

\subsection{Probability estimation\label{s42}}

The discrepancy measure (\ref{e9}) is appropriate for procedures that predict
a class identity and the corresponding contrast tree attempts to identify
$\mathbf{x}$ - values associated with high levels of misclassification. Some
procedures such as gradient boosting return estimated class probabilities at
each $\mathbf{x}$ which are then thresholded to predict class identities. In
this case the probability estimate contains information concerning expected
classification accuracy. The closer the respective class probabilities are to
each other the higher is the likelihood of misclassification. This shifts the
issue from classification accuracy to probability estimation accuracy which
can be assessed with a contrast tree.

For binary classification a natural discrepancy for probability estimation is%
\begin{equation}
d_{m}=\frac{1}{N_{m}}\left\vert \sum_{i\in R_{m}}(y_{i}-z_{i})\right\vert
\label{e10}%
\end{equation}
where $y\in\{0,1\}$ is the binary outcome variable and $0<z<1$ is its
predicted probability $\widehat{\Pr}(y=1)$. This (\ref{e10}) measures the
difference between the empirical probability of $y=1$ in region $R_{m}$ and
the corresponding average probability prediction $z$ in that region. The
contrast tree was built on the test data set with the gradient boosting
probability estimates based on the training data.%

%TCIMACRO{\FRAME{ftbpFU}{3.7429in}{3.2759in}{0pt}{\Qcb{Census income data.
%Upper frame: fraction of positive observations (blue) and mean probability
%prediction (red) for probability contrast tree regions. Lower frame:
%observation count in each region.}}{\Qlb{fig2}}{adultgbldiag.eps}%
%{\special{ language "Scientific Word";  type "GRAPHIC";  display "USEDEF";
%valid_file "F";  width 3.7429in;  height 3.2759in;  depth 0pt;
%original-width 7.9952in;  original-height 10.5031in;  cropleft "0";
%croptop "1";  cropright "1";  cropbottom "0";
%filename '../adultGBLdiag.eps';file-properties "XNPEU";}} }%
%BeginExpansion
\begin{figure}
[ptb]
\begin{center}
\includegraphics[
height=3.2759in,
width=3.7429in
]%
{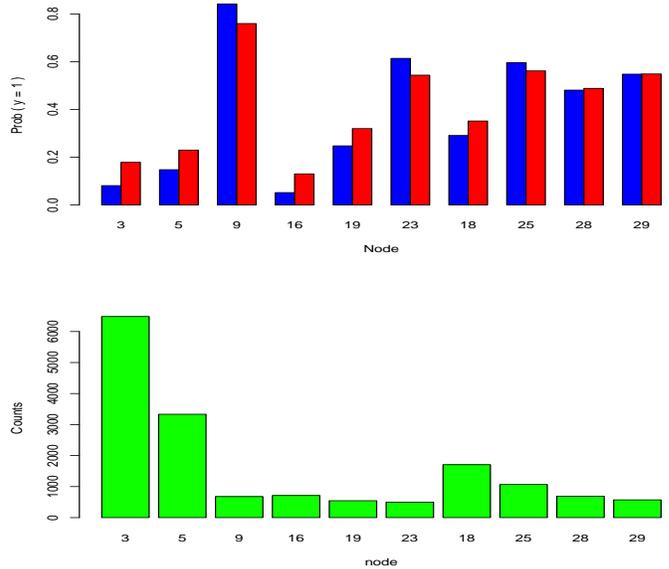}%
\caption{Census income data. Upper frame: fraction of positive observations
(blue) and mean probability prediction (red) for probability contrast tree
regions. Lower frame: observation count in each region.}%
\label{fig2}%
\end{center}
\end{figure}
%EndExpansion

The top frame of Fig. \ref{fig2} shows the empirical probability $y=1$ (blue)
and the average gradient boosting prediction $z$ (red) within each region of
the resulting contrast tree. The bottom frame shows the number of counts in
each corresponding region. One sees a general trend of over-smoothing. The
large probabilities are being under-estimated whereas the smaller ones are
substantially over-estimated by the gradient boosting procedure. In node $3$
containing 40\% of the observations the empirical $\Pr(y=1)$ is $0.037$
whereas the mean of the predictions is $\bar{z}=$ $0.18$. In node $16$ the
empirical probability is $0.05$ while the gradient boosting mean prediction in
that region is $0.17$. In node $9$ gradient boosting is under-estimating the
actual probability $\Pr(y=1)=0.84$ with $\bar{z}=$ $0.73$. As above these
regions are defined by simple rules based on the values of a few predictor variables.

A convenient way to summarize the\ overall results of a contrast tree is
through its corresponding lack-of-fit contrast curve. A discrepancy value is
assigned to each observation as being that of the corresponding region
containing it. Observations are then ranked on their assigned discrepancies.
For each one the mean discrepancy of those observations with greater than or
equal rank is computed. This is plotted on the vertical axis versus normalized
rank (fraction) on the horizontal axis. The left most point on each curve thus
represents the discrepancy value of the largest discrepancy region of its
corresponding tree. The right most point gives the discrepancy averaged over
all observations. Intermediate points give average discrepancy over the
highest discrepancy observations containing the corresponding fraction.%

%TCIMACRO{\FRAME{ftbpFU}{3.5137in}{3.339in}{0pt}{\Qcb{Census income data.
%Lack-of-fit contrast curves comparing probability $y=1$ estimates by logistic
%gradient boosting (black), random forests (blue), squared-error gradient
%boosting (red) and contrast boosting (green).}}{\Qlb{fig2.5}}{adultgbrf.eps}%
%{\special{ language "Scientific Word";  type "GRAPHIC";  display "USEDEF";
%valid_file "F";  width 3.5137in;  height 3.339in;  depth 0pt;
%original-width 7.9952in;  original-height 10.5031in;  cropleft "0";
%croptop "1";  cropright "1";  cropbottom "0";
%filename '../adultgbrf.eps';file-properties "XNPEU";}} }%
%BeginExpansion
\begin{figure}
[ptb]
\begin{center}
\includegraphics[
height=3.339in,
width=3.5137in
]%
{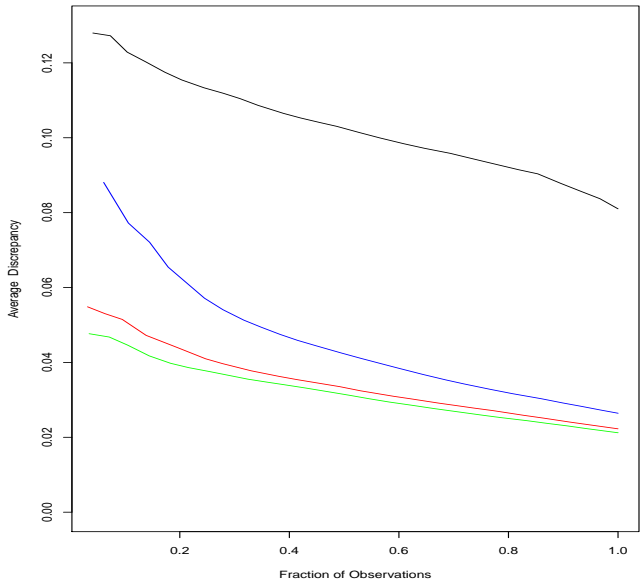}%
\caption{Census income data. Lack-of-fit contrast curves comparing probability
$y=1$ estimates by logistic gradient boosting (black), random forests (blue),
squared-error gradient boosting (red) and contrast boosting (green).}%
\label{fig2.5}%
\end{center}
\end{figure}
%EndExpansion

The black curve in Fig. \ref{fig2.5} \ shows the lack-of-fit contrast curve
for the gradient boosting estimates based on a 50 node contrast tree. Its
error in estimated probability averaged over all test set predictions is seen
to be $0.081$ (extreme right). The error corresponding to the largest
discrepancy region (extreme left) is $0.128$. The blue curve is the
corresponding lack-of-fit curve for random forest probability prediction
(Breiman 2001). Its average error is half of that for gradient boosting and
its worst error is 30\% less.

The contrast tree represented in Fig. \ref{fig2} suggests that the problem
with the gradient boosting procedure here is over-smoothng. It is failing to
accurately estimate the extreme probability values. Gradient boosting for
binary probability estimation generally uses a Bernoulli log--likelihood loss
function based on a logistic distribution. The logistic transform has the
effect of suppressing the influence of extreme probabilities. Random forests
are based on squared--error loss. This suggests that using a different loss
function with gradient boosting for this problem may improve performance,
especially at the extreme values.

The red curve in Fig. \ref{fig2.5} \ shows the corresponding lack-of-fit
contrast curve for gradient boosting probability estimates using
squared--error loss. This change in loss criterion has dramatically improved
accuracy of gradient boosting estimates. Both its average and maximum
discrepancies are seen to be at least three times smaller than those using the
logistic regression based loss criterion.

The green contrast curve in Fig. \ref{fig2.5} represents results of contrast
\emph{boosting} applied to the output of squared-error loss gradient boosting
as discussed in Section \ref{s5.1}. It is seen to provide only little
improvement here. The quantile regression example provided in the Appendix
however shows substantial improvement when contrast boosting is applied to the
gradient boosting output.

Standard errors on the results shown in Fig. \ref{fig2.5} can be computed
using bootstrap resampling (Efron 1979). Standard errors on the left most
points, representing the highest discrepancy regions of the respective curves,
are $0.0089$, $0.016$, $0.012$, and $0.012$ (top to bottom). For the right
most points on the curves, representing average tree discrepancy, the
corresponding errors are $0.0037$, $0.0026$, $0.0026$ and $0.0026$. Errors
corresponding to intermediate points on each curve are between those of its
two extremes

\begin{center}
\textbf{Table 1}

Classification error rates corresponding to the several probability estimation methods.

$%
\begin{array}
[c]{ll}%
\text{Method} & \text{Error rate}\\
\text{Gradient Boosting -- logistic loss} & 13.0\%\\
\text{Gradient Boosting -- squared-error loss} & 12.9\%\\
\text{Random Forest} & 13.6\%\\
\text{Contrast Boosting} & 12.8\%
\end{array}
$
\end{center}

Table 1 shows classification error rate for each of the methods considered.
They are all seen to be very similar. This illustrates that classification
prediction accuracy can be a very poor proxy for probability estimation
accuracy. In this case the over--smoothing of probability estimates caused by
using the logistic log--likelihood does not change many class assignments.
However, in applications that require high precision accurate estimation of
extreme probabilities is more important.

\subsection{Conditional distributions\label{s44}}

Here we consider the case in which both $y$ and $z$ are considered to be
random variables independently drawn from respective distributions
$p_{y}(y\,|\,\mathbf{x})$ and $p_{z}(z\,|\,\mathbf{x})$. Interest is in
contrasting these two distributions as functions of $\mathbf{x}$. Specifically
we wish to uncover regions of $\mathbf{x}$ - space where the distributions
most differ. For this we use contrast trees (Section \ref{s2}) with
discrepancy measure (\ref{e8})$.$

A well known way to approximate $p_{y}(y\,|\,\mathbf{x})$ under the assumption
of homoskedasticity is through the residual bootstrap (Efron and Tibshirani
1994). One obtains a location estimate such as the conditional median $\hat
{m}(y\,|\,\mathbf{x})$ and forms the data residuals $r_{i}=y_{i}-\hat
{m}(y\,|\,\mathbf{x}_{i})$ for each observation $1\leq i\leq N$. Under the
assumption that the conditional distribution of $r$, $p_{r}(r\,|\mathbf{\,x}%
)$, is independent of $\mathbf{x}$ (homoskedasticity) one can draw random
samples from $p_{y}(y\,|\,\mathbf{x}_{i})$ as $y_{i}=\hat{m}(y\,|\,\mathbf{x}%
_{i})+r_{\pi(i)}$ where $\pi(i)$ is random permutation of the integers
$i\in\,[1,N]$. These samples can then be used to derive various regression
statistics of interest.

A fundamental ingredient for the validity of residual bootstrap approach is
the homoskedasticity assumption. Here we test this on the online news
popularity data set (Fernandes, Vinagre and Cortez, 2015) also available from
the Irvine Machine Learning Data Repository. It summarizes a heterogeneous set
of features about articles published by Mashable web site over a period of two
years. The goal is to predict the number of shares $y$ in social networks
(popularity). There are $N=39797$ observations (articles). Associated with
each are $p=59$ attributes to be used as predictor variables $\mathbf{x}$.
These are described at the download web site. Gradient boosting was used to
estimate the median function $\hat{m}(y\,|\,\mathbf{x})$, and $\{z_{i}%
\}_{i=1}^{N}$ was taken as a corresponding residual bootstrap sample to be
contrasted with $y$.%

%TCIMACRO{\FRAME{ftbpFU}{3.7222in}{3.0779in}{0pt}{\Qcb{ QQ--plots of $y$ versus
%parametric bootstrap $z$ distributions for the nine highest discrepancy
%regions of a 50 node contrast tree using online news popularity data. The red
%line represents equality.}}{\Qlb{fig4}}{newsdistshare.eps}%
%{\special{ language "Scientific Word";  type "GRAPHIC";  display "USEDEF";
%valid_file "F";  width 3.7222in;  height 3.0779in;  depth 0pt;
%original-width 7.9952in;  original-height 10.5031in;  cropleft "0";
%croptop "1";  cropright "1";  cropbottom "0";
%filename 'newsdistshare.eps';file-properties "XNPEU";}} }%
%BeginExpansion
\begin{figure}
[ptb]
\begin{center}
\includegraphics[
height=3.0779in,
width=3.7222in
]%
{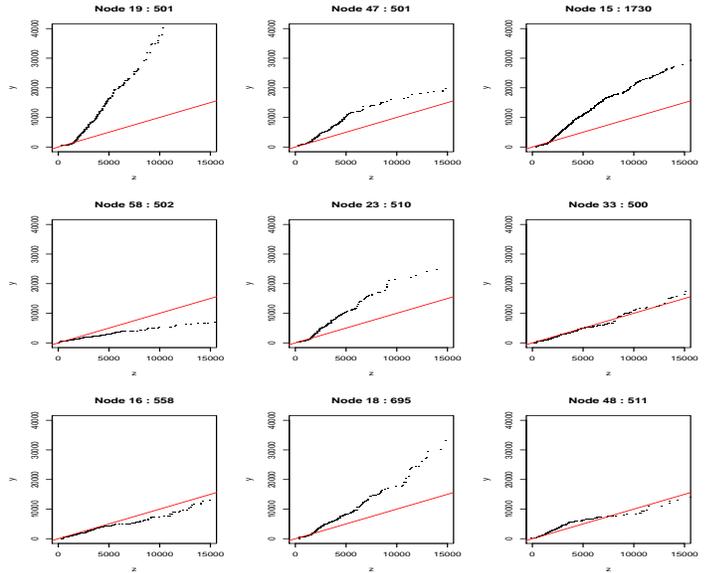}%
\caption{ QQ--plots of $y$ versus parametric bootstrap $z$ distributions for
the nine highest discrepancy regions of a 50 node contrast tree using online
news popularity data. The red line represents equality.}%
\label{fig4}%
\end{center}
\end{figure}
%EndExpansion

Figure \ref{fig4} shows quantle-quantile (QQ)-plots of $y$ versus $z$ for the
nine highest discrepancy regions of a 50 node contrast tree. The red line
represents equality. One sees that there are $\mathbf{x}$ - values (regions)
where the distribution of $y$ is very different from its residual bootstrap
approximation $z$; homoskedasticity is rather strongly violated. The average
discrepancy (\ref{e8}) over all 50 regions is $0.19$.%

%TCIMACRO{\FRAME{ftbpFU}{3.5449in}{2.7354in}{0pt}{\Qcb{ QQ--plots of $\log
%_{10}(y)$ versus corresponding parametric bootstrap $z$ distributions for the
%nine highest discrepancy regions of a 50 node contrast tree using online news
%popularity data. The red line represents equality.}}{\Qlb{fig5}}%
%{newsdistlog.eps}{\special{ language "Scientific Word";  type "GRAPHIC";
%display "USEDEF";  valid_file "F";  width 3.5449in;  height 2.7354in;
%depth 0pt;  original-width 7.9952in;  original-height 10.5031in;
%cropleft "0";  croptop "1";  cropright "1";  cropbottom "0";
%filename 'newsdistlog.eps';file-properties "XNPEU";}} }%
%BeginExpansion
\begin{figure}
[ptb]
\begin{center}
\includegraphics[
height=2.7354in,
width=3.5449in
]%
{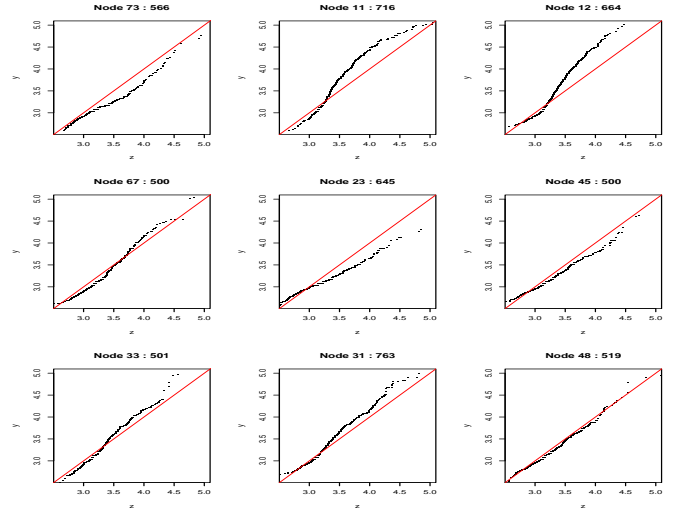}%
\caption{ QQ--plots of $\log_{10}(y)$ versus corresponding parametric
bootstrap $z$ distributions for the nine highest discrepancy regions of a 50
node contrast tree using online news popularity data. The red line represents
equality.}%
\label{fig5}%
\end{center}
\end{figure}
%EndExpansion

The outcome variable $y$ (number of shares) is strictly positive and its
marginal distribution is highly skewed toward larger values. In such
situations it is common to model its logarithm. Figure \ref{fig5} shows the
corresponding results for contrasting the distribution of $\log_{10}(y)$ with
its residual bootstrap counterpart. Homoskedasticity appears to more closely
hold on the logarithm scale but there are still regions of $\mathbf{x}$ -
space where the approximation is not good. Here the average discrepancy
(\ref{e8}) over all 50 regions is $0.13$. A null distribution for average
discrepancy under the hypothesis of homoskedasticity can be obtained by
repeatedly contrasting pairs of randomly generated $\log_{10}(y)$ residual
bootstrap distributions. Based on 50 replications, this distribution had a
mean of $0.078$ with a standard deviation of $0.003$.

\section{Distribution boosting -- simulated data\label{tbs}}

The notion of distribution boosting (Section \ref{s5.2}) is sufficiently
unusual that we first illustrate it on simulated data where the estimates
$\hat{p}_{y}(y\,|\,\mathbf{x})$ can be compared to the true data generating
distributions $p_{y}(y\,|\,\mathbf{x})$. Distribution boosting applied to the
online news popularity data described in Section \ref{s44} is presented in the Appendix.

\subsubsection{Data\label{simdat}}

There are $N=25000$ training observations each with a set\ of $p=10$ predictor
variables $\mathbf{x}_{i}$ randomly generated from a standard normal
distribution. The outcome variable $y\,|\,\mathbf{x}$ is generated from a
transformed \emph{asymmetric} logistic distribution (Friedman 2018)
\begin{equation}
y=h\,(f(\mathbf{x})+\eta(\mathbf{x})) \label{e36}%
\end{equation}
with
\[
\text{\ }%
\begin{array}
[c]{l}%
\eta(\mathbf{x})=-s_{l}(\mathbf{x})\cdot|\,\varepsilon\,|\text{,
\ \ }prob=s_{l}(\mathbf{x})/(s_{l}(\mathbf{x})+s_{u}(\mathbf{x}))\\
\eta(\mathbf{x})=+s_{u}(\mathbf{x})\cdot|\,\varepsilon\,|\text{,\ \ }%
prob=s_{u}(\mathbf{x})/(s_{l}(\mathbf{x})+s_{u}(\mathbf{x}))\text{.}%
\end{array}
\]
Here $\varepsilon$ is a standard logistic random variable, and the
transformation $h(z)$ is
\begin{equation}
h(z)=sign(z)\,(0.5\,|\,z\,|\,+1.5\,\,z^{2})\text{.} \label{e19}%
\end{equation}

The untransformed mode $f(\mathbf{x})$ and lower/upper scales $s_{l}%
(\mathbf{x})$\thinspace/$\,s_{u}(\mathbf{x})$ are each different functions of
the ten predictor variables $\mathbf{x}$. The simulated mode function is taken
to be%

\begin{equation}
f(\mathbf{x})=\sum_{j=1}^{10}c_{j}\,B_{j}(x_{j})\,/\,std_{x_{j}}(B_{j}%
(x_{j})\,) \label{e20}%
\end{equation}
with the value of each coefficient $c_{j}$ being randomly drawn from a
standard normal distribution. Each basis function takes the form
\begin{equation}
B_{j}(x_{j})=sign(x_{j})\,|\,x_{j}\,|^{r_{j}} \label{e21}%
\end{equation}
with each exponent $r_{j}$\ being separately drawn from a uniform distribution
$r_{j}\sim U(0,2)$. The denominator in each term of (\ref{e20}) prevents the
suppression of the influence of highly nonlinear terms in defining
$f(\mathbf{x})$.

The scale functions are taken to be $s_{l}(\mathbf{x})=0.2+\exp\left(
t_{l}(\mathbf{x})\right)  $ and $s_{u}(\mathbf{x})=0.2+\exp(t_{u}%
(\mathbf{x}))$ where the log--scale functions $t_{l}(\mathbf{x})$ and
$t_{u}(\mathbf{x})$ are constructed in the same manner as (\ref{e20})
(\ref{e21}) but with different randomly drawn values for the $20$ parameters
$\{c_{j},r_{j}\}_{1}^{10}$ producing different functions of $\mathbf{x}$. The
average pair-wise absolute correlation between the three functions is $0.18$.
The overall resulting distribution $p(y\,|\,\mathbf{x})$ (\ref{e36}%
--\ref{e21}) has location, scale, asymmetry, and shape being highly dependent
on the joint values of the predictors $\mathbf{x}$ in a complex and unrelated way.

\subsubsection{Conditional distribution estimation\label{s63}}

Distribution boosting is applied to this simulated data to estimate its
distribution $p_{y}(y\,|\,\mathbf{x})$ as a function of $\mathbf{x}$. For each
observation the contrasting random variable $z$ is taken to be independently
generated from a standard normal distribution, $z\,|\,\mathbf{x}\sim
N(0,1)\,$, independent of $\mathbf{x}$. The goal is to produce an estimated
transformation of $z$, $\hat{y}=\hat{g}_{\mathbf{x}}(z)$, at each $\mathbf{x}$
such that $p_{\hat{y}}(\hat{y}\,|\,\mathbf{x})=$ $p_{y}(y\,|\,\mathbf{x})$. To
the extent the estimate $\hat{g}_{\mathbf{x}}(z)$ accurately reflects the true
transformation function $g_{\mathbf{x}}(z)$ at each $\mathbf{x}$ one can apply
it to a sample of standard normal random numbers to produce a sample drawn
from the distribution $p_{y}(y\,|\,\mathbf{x})$. This sample can then be used
to plot that distribution or compute the value of any of its properties.%

%TCIMACRO{\FRAME{ftbpFU}{3.768in}{2.8608in}{0pt}{\Qcb{Test data discrepancy
%averaged over the terminal nodes (regions) of successive contrast trees for
%the first and then every tenth iteration for 400 iterations of distribution
%boosting on simulated training data. The solid red curve is a running median
%smooth.}}{\Qlb{fig10}}{simtestdisc.eps}{\special{ language "Scientific Word";
%type "GRAPHIC";  display "USEDEF";  valid_file "F";  width 3.768in;
%height 2.8608in;  depth 0pt;  original-width 7.9952in;
%original-height 10.5031in;  cropleft "0";  croptop "1";  cropright "1";
%cropbottom "0";  filename 'simtestdisc.eps';file-properties "XNPEU";}} }%
%BeginExpansion
\begin{figure}
[ptb]
\begin{center}
\includegraphics[
height=2.8608in,
width=3.768in
]%
{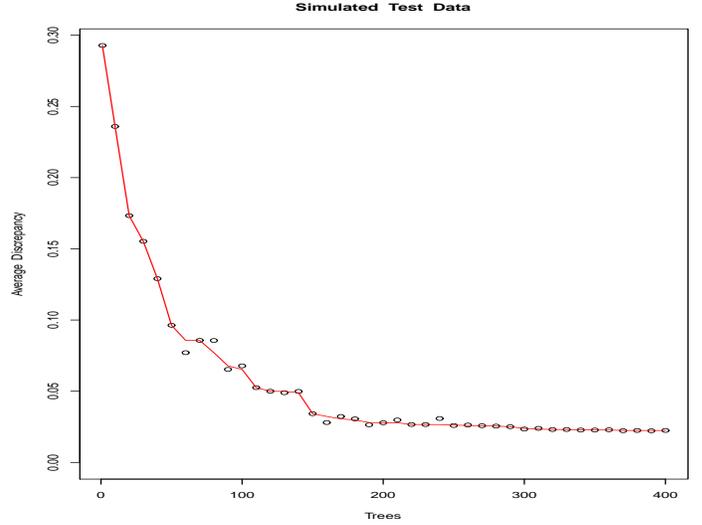}%
\caption{Test data discrepancy averaged over the terminal nodes (regions) of
successive contrast trees for the first and then every tenth iteration for 400
iterations of distribution boosting on simulated training data. The solid red
curve is a running median smooth.}%
\label{fig10}%
\end{center}
\end{figure}
%EndExpansion

Figure \ref{fig10} plots the average terminal node discrepancy (\ref{e8}) for
400 iterations of distribution boosting applied to the training data, as
evaluated on a $25000$ observation independent \textquotedblleft
test\textquotedblright\ data set generated from the same joint $(\mathbf{x}%
,y)$ - distribution (\ref{e36} -- \ref{e21}). Results are shown for the first
and then every tenth successive tree. The test set discrepancy is seen to
generally decrease with increasing number of trees. There is a diminishing
return after about 200 iterations (trees).

Note that with contrast boosting average tree discrepancy on test or training
data does not necessarily decrease monotonically with successive iterations
(trees). Each contrast tree represents a greedy solution to a non convex
optimization with multiple local optima. As a consequence the inclusion of an
additional tree can, and often does, increase average discrepancy of the
current ensemble. Boosting is continued as long as there is a general downward
trend in average tree discrepancy.

\begin{figure}
[ptb]
\begin{center}
\includegraphics[
height=3.0338in,
width=3.6573in
]%
{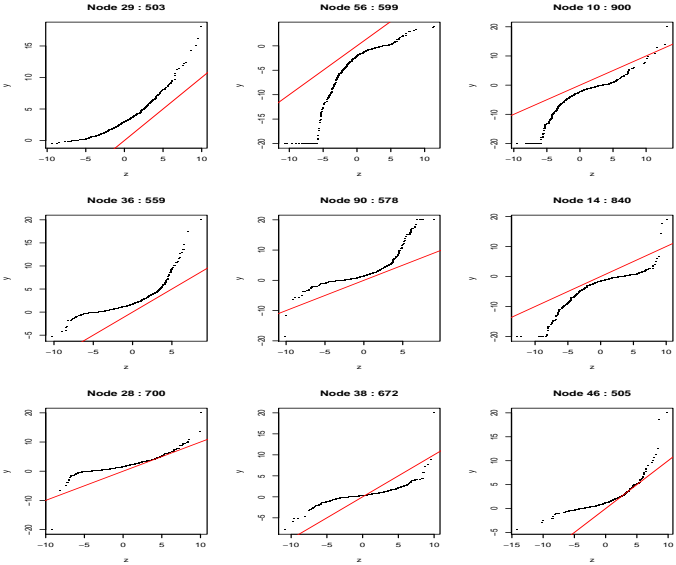}%
\caption{QQ--plots of $y$ versus $z$ (standard normal) for the nine highest
discrepancy regions of a 10 node contrast tree on the simulated test data set.
The red lines represent equality.}%
\label{fig11}%
\end{center}
\end{figure}

Lack-of-fit to the data of any model for the distribution $p_{y}%
(y\,|\,\mathbf{x})$ can be can be assessed by contrasting $y$ with a sample
drawn from that distribution. Figure \ref{fig11} shows QQ--plots of $y$ versus
initial $z$ (everywhere standard normal) for the nine highest discrepancy
regions of a 10 node tree contrasting the two quantities on the test data set.
The red lines represent equality. One sees that $p_{y}(y\,|\,\mathbf{x})$ is
here far from being everywhere standard normal.%

%TCIMACRO{\FRAME{ftbpFU}{3.6573in}{3.0338in}{0pt}{\Qcb{QQ--plots of $y$ versus
%$z$ (standard normal) for the nine highest discrepancy regions of a 10 node
%contrast tree on the simulated test data set. The red lines represent
%equality.}}{\Qlb{fig11}}{simgoforig.eps}%
%{\special{ language "Scientific Word";  type "GRAPHIC";  display "USEDEF";
%valid_file "F";  width 3.6573in;  height 3.0338in;  depth 0pt;
%original-width 7.9952in;  original-height 10.5031in;  cropleft "0";
%croptop "1";  cropright "1";  cropbottom "0";
%filename 'simgoforig.eps';file-properties "XNPEU";}} }%
%BeginExpansion
%EndExpansion

\begin{figure}
[ptb]
\begin{center}
\includegraphics[
height=3.0234in,
width=3.5423in
]%
{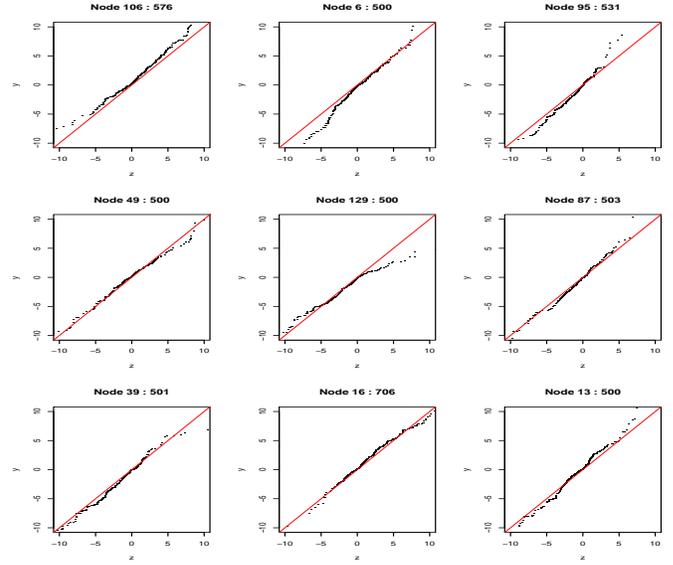}%
\caption{QQ--plots of $y$ versus $\hat{y}=\hat{g}_{\mathbf{x}}(z)$ for the
nine highest discrepancy regions of a 10 node contrast tree on the simulated
test data set. The red lines represent equality.}%
\label{fig11.5}%
\end{center}
\end{figure}

For the distribution boosted model $\hat{y}=\hat{g}_{\mathbf{x}}(z)$
lack-of-fit can be assessed by contrasting the distributions of $y$ and
$\hat{y}$ with a contrast tree using the test data set. Figure \ref{fig11.5}
shows QQ--plots of $y$ versus $\hat{y}$ for the nine highest discrepancy
regions of a 10 node tree contrasting the two quantities on the test data set.
The red lines represent equality. The transformation $\hat{g}_{\mathbf{x}}(z)$
at each separate $\mathbf{x}$ - value was evaluated using the $400$ tree model
built on the training data. The nine highest discrepancy regions shown in Fig.
\ref{fig11.5} together cover 27\% of the data. They show that while the
transformation model fits most of the test data quite well, it is not
everywhere perfect. There are minor departures between the two distributions
in some small regions. However these discrepancies appear in sparse tails
where QQ--plots themselves tend to be unstable.%

%TCIMACRO{\FRAME{ftbpFU}{3.5423in}{3.0234in}{0pt}{\Qcb{QQ--plots of $y$ versus
%$\hat{y}=\hat{g}_{\mathbf{x}}(z)$ for the nine highest discrepancy regions of
%a 10 node contrast tree on the simulated test data set. The red lines
%represent equality.}}{\Qlb{fig11.5}}{simgofsoln.eps}%
%{\special{ language "Scientific Word";  type "GRAPHIC";  display "USEDEF";
%valid_file "F";  width 3.5423in;  height 3.0234in;  depth 0pt;
%original-width 7.9952in;  original-height 10.5031in;  cropleft "0";
%croptop "1";  cropright "1";  cropbottom "0";
%filename 'simgofsoln.eps';file-properties "XNPEU";}} }%
%BeginExpansion
%EndExpansion

A measure of the difference between the estimated and true CDFs can be defined
as
\begin{equation}
AAE=\frac{1}{100}\sum_{j=1}^{100}|\,CDF(u_{j})-\widehat{CDF}\,(u_{j})\,|
\label{e22}%
\end{equation}
where $CDF$ is the true cumulative distribution of $y\,|\,\mathbf{x}$ computed
form \ (\ref{e36}--\ref{e21}) and $\widehat{CDF}\,$\ is the corresponding
estimate from the distribution boosting model. The $100$ evaluation points
$\{u_{j}\}_{1}^{100}$ are a uniform grid between the $0.001$ and $0.999$
quantiles of the true distribution $CDF$.%

%TCIMACRO{\FRAME{ftbpFU}{3.7507in}{3.1488in}{0pt}{\Qcb{Upper left: CDF
%estimation error distribution for simulated data. Upper right: estimated
%(black) and true (red) CDFs for observation with median error. Lower:
%corresponding plots for 75\% and 90\% decile errors.}}{\Qlb{fig13}%
%}{simhistplots.eps}{\special{ language "Scientific Word";  type "GRAPHIC";
%display "USEDEF";  valid_file "F";  width 3.7507in;  height 3.1488in;
%depth 0pt;  original-width 7.9952in;  original-height 10.5031in;
%cropleft "0";  croptop "1";  cropright "1";  cropbottom "0";
%filename 'simhistplots.eps';file-properties "XNPEU";}} }%
%BeginExpansion
\begin{figure}
[ptb]
\begin{center}
\includegraphics[
height=3.1488in,
width=3.7507in
]%
{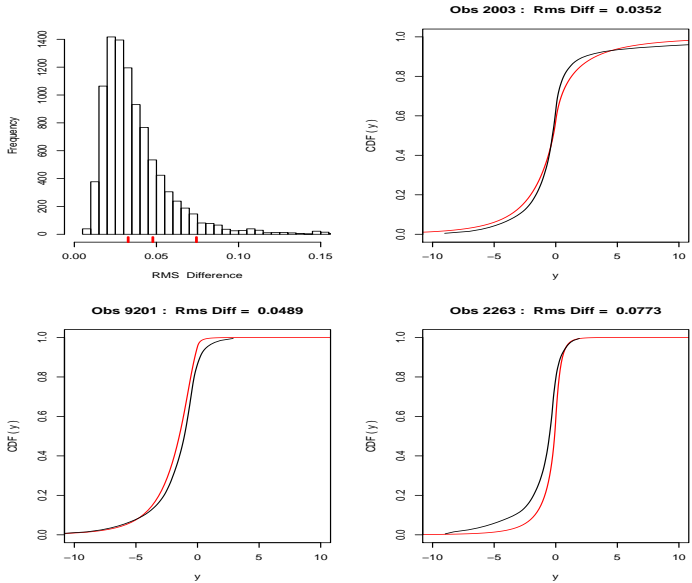}%
\caption{Upper left: CDF estimation error distribution for simulated data.
Upper right: estimated (black) and true (red) CDFs for observation with median
error. Lower: corresponding plots for 75\% and 90\% decile errors.}%
\label{fig13}%
\end{center}
\end{figure}
%EndExpansion

Figure \ref{fig13} summarizes the overall accuracy of the distribution
boosting model. The upper left frame shows a histogram of the distribution of
(\ref{e22}) for observations in the test data set. The 50, 75 and 90
percentiles of this distribution are respectively $0.0352$, $0.0489$ and
$0.0773$ indicated by the red marks. The remaining plots show estimated
(black) and true (red) distributions for three observations with (\ref{e22})
closest to these respective percentiles. Thus 50\% of the estimated
distributions are closer to the truth than that shown in the upper right
frame. Seventy five percent are closer than that shown in the lower left
frame, and 90\% are closer than that seen in the lower right frame.

Distribution boosting produces an estimate for the full distribution of
$y\,|\,\mathbf{x}$ by providing a function $\hat{g}_{\mathbf{x}}(z)$ that
transforms a random variable $z$ with a known distribution $p_{z}%
(z\,|\,\mathbf{x})$ to the estimated distribution $\hat{p}_{y}%
(y\,|\,\mathbf{x})$. One can then easily compute any statistic $\hat
{S}(\mathbf{x})=S[$ $\hat{p}_{y}(y\,|\,\mathbf{x})]$, which can be used as an
estimate for the value of the corresponding quantity $S(\mathbf{x})=S[$
$p_{y}(y\,|\,\mathbf{x})]$ on the actual distribution. For some quantities
$S(\mathbf{x})$ an alternative is to directly \ estimate them by minimizing
empirical prediction risk based on an appropriate loss function%
\begin{equation}
\hat{S}(\mathbf{x})=\arg\min_{f\in\Im}\frac{1}{N}\sum_{i=1}^{N}L(y_{i}%
,f(\mathbf{x}_{i})) \label{e24}%
\end{equation}
were $\Im$ is the function class associated with the learning method. Here we
compare distribution boosting (DB) estimates of the quartiles $Q_{p}%
(\mathbf{x)}$, $p\in\lbrack0.25,0.5,0.75]$, with those of gradient boosting
quantile regression (GB), which uses loss
\begin{equation}
L_{p}(y,z)=(1-p)\,(z-y)_{+}+p\,(y-z)_{+}\text{,} \label{e24.5}%
\end{equation}
on the simulated data set where the truth is known.%

%TCIMACRO{\FRAME{ftbpFU}{3.6884in}{2.4613in}{0pt}{\Qcb{Predicted versus true
%values for the three quartiles as functions of $\mathbf{x}$ (columns) for
%gradient boosting quantile regression (upper row) and distribution boosting
%(lower row) on the simulated data. The red lines represent a running median
%smooth and the blue lines show equality.}}{\Qlb{fig14}}{simcompquant.eps}%
%{\special{ language "Scientific Word";  type "GRAPHIC";  display "USEDEF";
%valid_file "F";  width 3.6884in;  height 2.4613in;  depth 0pt;
%original-width 7.9952in;  original-height 10.5031in;  cropleft "0";
%croptop "1";  cropright "1";  cropbottom "0";
%filename 'simcompquant.eps';file-properties "XNPEU";}} }%
%BeginExpansion
\begin{figure}
[ptb]
\begin{center}
\includegraphics[
height=2.4613in,
width=3.6884in
]%
{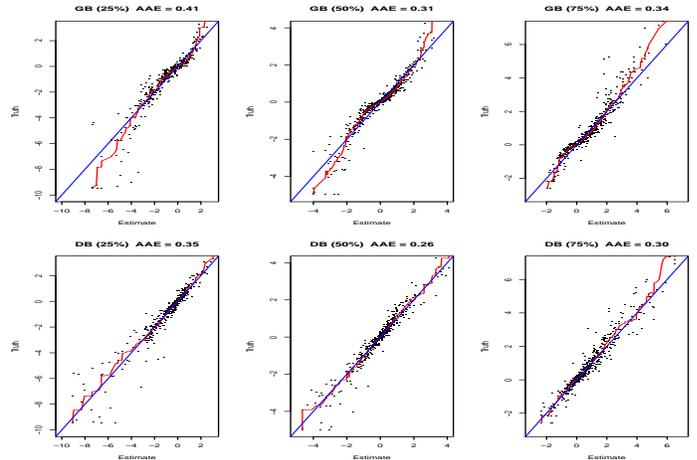}%
\caption{Predicted versus true values for the three quartiles as functions of
$\mathbf{x}$ (columns) for gradient boosting quantile regression (upper row)
and distribution boosting (lower row) on the simulated data. The red lines
represent a running median smooth and the blue lines show equality.}%
\label{fig14}%
\end{center}
\end{figure}
%EndExpansion

Figure \ref{fig14} shows true versus predicted values for each of the two
methods (rows) on the three quartiles (columns). The red lines represent a
running median smooth and the blue lines show equality. The average absolute
error $AAE$ associated with each of these plots is%
\begin{equation}
AAE(h,v)=mean(|\,h-v\,|)/mean(|\,v-median(v)\,|) \label{e25}%
\end{equation}
where $h$ is the quantity plotted on the horizontal and $v$ the vertical axes.
The quantile values derived from the estimates of the full distribution
(bottom row) are here seen to be somewhat more accurate than those obtained
from gradient boosting quantile regression (top row).

With quantile regression each quantile is estimated separately without regard
to estimates of other quantiles. Distribution boosting quantile estimates are
all derived from a common probability distribution and thus have constraints
imposed among them. For example, two quantile estimates have the property
$\hat{Q}_{p}(\mathbf{x})<\hat{Q}_{p^{\prime}}(\mathbf{x})$ for all $p<$
$p^{\prime}$ at any $\mathbf{x}$. These constraints can improve accuracy
especially when the quantile estimates are being used to compute quantities
derived from them.

There is an additional advantage of computing quantities such as means or
quantiles\ from the estimated conditional distributions $\hat{S}%
(\mathbf{x})=S[$ $\hat{p}_{y}(y\,|\,\mathbf{x})]$. As noted in Section
\ref{s3}, distribution contrast trees can be constructed in the presence of
arbitrary censoring or truncation. This extends to contrast boosted
distribution estimates $\hat{p}_{y}(y\,|\,\mathbf{x})$ and any quantities
derived from them. This in turn allows application to ordinal regression which
can be considered a special case of interval censoring (Friedman 2018).

\section{Related work\label{s8}}

Regression trees have a long history in Statistics and Machine Learning. Since
their first introduction (Morgan and Songquist 1963) many proposed
modifications have been introduced to increase accuracy and extend
applicability. See Loh (2014) for a nice survey. More recent extensions
include Mediboost (Valdes \emph{et al} 2016) and the Additive Tree (Luna
\emph{et al} 2019). All of these proposals are focused towards estimating the
properties of a single outcome variable. There seems to have been little to no
work related to applications involving contrasting two variables.

Friedman and Fisher (1999) proposed using tree style recursive partitioning
strategies to identify interpretable regions in $\mathbf{x}$ - space within
which the mean of a single outcome $y$ was relatively large (\textquotedblleft
hot spots\textquotedblright). With a similar goal Buja and Lee (2001) proposed
using ordinary regression trees with a splitting criterion based on the
maximum of the two daughter node means.

Classification tree boosting was proposed by Freund and Schapire (1997).
Extension to regression trees was developed by Friedman (2001). Since then
there has been considerable research attempting to improve accuracy and
extending its scope. See Mayr \emph{et al} (2014) for a good summary.

Although boosted contrast trees have not been previously proposed they are
generally appropriate for the same types of applications as gradient boosted
regression trees, such as classification, regression, and quantile regression.
They can be beneficial in applications where a contrast tree indicates
lack-of-fit of a model produced by some estimation method. In such situations
applying contrast boosting to the model predictions often provides improvement
in accuracy.

Tree ensembles have also been applied to nonparametric conditional
distribution estimation. Meinshausen (2006) used classical random forests to
define local neighborhoods in $\mathbf{x}$ - space. The empirical conditional
distribution in each such defined local region around a prediction point
$\mathbf{x}$ is taken as the corresponding conditional distribution estimate
at $\mathbf{x}$. Athey, Tibshirani and Wagner (2019) noted that since the
regression trees used by random forests are designed to detect only mean
differences the resulting neighborhoods will fail to adequately capture
distributions for which higher moments are not generally functions of the
mean. They proposed modified tree building strategies based on gradient
boosting ideas to customize random forest tree construction for specific
applications including quantile regression.

Boosted regression trees have been used as components in procedures for
parametric fitting of conditional distributions and transformations. A
parametric form for the conditional distribution or transformation is
hypothesized and the parameters as functions of $\mathbf{x}$ are estimated by
regression tree gradient boosting using negative log--likelihood as the
prediction risk. See for example Mayr et al (2012), Friedman (2018), Pratola
\emph{et al} (2019), Hothorn (2019) and Mukhopadlhyay \& Wang (2019). Some
differences between these previous methods and the corresponding approaches
proposed here include use of contrast rather than regression trees, and no
parametric assumptions.

The principal benefit of the contrast tree based procedures is a lack-of-fit
measure. As seen in Table 1 of Section \ref{s42}, and in the Appendix, values
of negative log--likelihoods or prediction risk need not reflect actual
lack-of-fit to the data. The values of their minima can depend upon other
unmeasured quantities. The goal of contrast trees as illustrated in this paper
is to provide such a measure. Contrast trees can be applied to assess
lack-of-fit of estimates produced by any method, including those mentioned
above. If discrepancies are detected, contrast boosting can be employed to
remedy them and thereby improve accuracy.

\section{Summary\label{s9}}

Contrast trees as described in Sections \ref{s2} and \ref{s3} are designed to
provide interpretable goodness-of-fit diagnostics for estimates of the
parameters of $p_{y}(y\,|\,\mathbf{x})$, or the full distribution. Examples
involving classification, probability estimation and conditional distribution
estimation were presented in Section \ref{s4}. A quantile regression example
is presented in the Appendix. Two--sample contrast trees for detecting
discrepancies between separate data sets are also described in the Appendix.

Boosting of contrast trees is a natural extension. Given an initial estimate
$\hat{z}(\mathbf{x})$ from any learning method a contrast tree can assess its
goodness or lack-of-fit to the data. If found lacking, the boosting strategy
attempts to improve the fit by successively modifying $\hat{z}(\mathbf{x})$ to
bring it closer to the data. The Appendix provides an example involving
quantile regression where this strategy substantially improved prediction accuracy.

Contrast boosting the full conditional distribution is illustrated on
simulated data in Section \ref{s63} and on actual data in the Appendix. Note
that the conditional distribution procedure of Section \ref{s5.2} can be
applied in the presence of arbitrarily censored or truncated data by employing
Turnbul's (1976) algorithm to compute CDFs and corresponding quantiles.

Contrast trees and boosting inherit all of the data analytic advantages of
classification and regression trees. These include categorical variables,
missing values, invariance to monotone transformations of the predictor
variables, resistance to irrelevant predictors, variable importance measures,
and few tuning parameters.

Important discussions with Trevor Hastie and Rob Tibshirani on the subject of
this work are gratefully acknowledged. An \emph{R} procedure implementing the
methods described herein is available.

\qquad

\begin{center}
{\huge Appendix}

\appendix{}
\end{center}

\section{Lack-of-fit estimation}

Here contrast tree lack-of-fit estimates are compared with known truth on
simulated data. There are $N=25000$ observations each with $p=10$ predictor
variables $\mathbf{x}$ randomly generated from a standard normal distribution.
The outcome $y$ is generated from a simple model%
\[
y=f(\mathbf{x})+s(\mathbf{x})\cdot\varepsilon
\]
with $\varepsilon$ a standard normal random variable. The location
$f(\mathbf{x})$ and scale $s(\mathbf{x})$ functions are given by (\ref{e20})
(\ref{e21}) with different randomly generated parameters. The correlation
between the two functions over the data is $cor(f(\mathbf{x}),s(\mathbf{x}))=$
$0.06$. The signal/noise is $IQR(f(\mathbf{x}))/(2\cdot med(s(\mathbf{x}%
)))=3$. The goal is to estimate the location function $f(\mathbf{x})$.%

%TCIMACRO{\FRAME{ftbpFU}{3.774in}{2.943in}{0pt}{\Qcb{ Lack-of-fit contrast
%curves on simulated data. Black: constant fit , purple: single CART tree,
%blue: linear model, violet: random forest, orange: squared-error and red:
%absolute loss gradient boosting, green: truth.}}{\Qlb{fig20}}{symsim.eps}%
%{\special{ language "Scientific Word";  type "GRAPHIC";  display "USEDEF";
%valid_file "F";  width 3.774in;  height 2.943in;  depth 0pt;
%original-width 7.9952in;  original-height 10.5031in;  cropleft "0";
%croptop "1";  cropright "1";  cropbottom "0";
%filename 'symsim.eps';file-properties "XNPEU";}} }%
%BeginExpansion
\begin{figure}
[ptb]
\begin{center}
\includegraphics[
height=2.943in,
width=3.774in
]%
{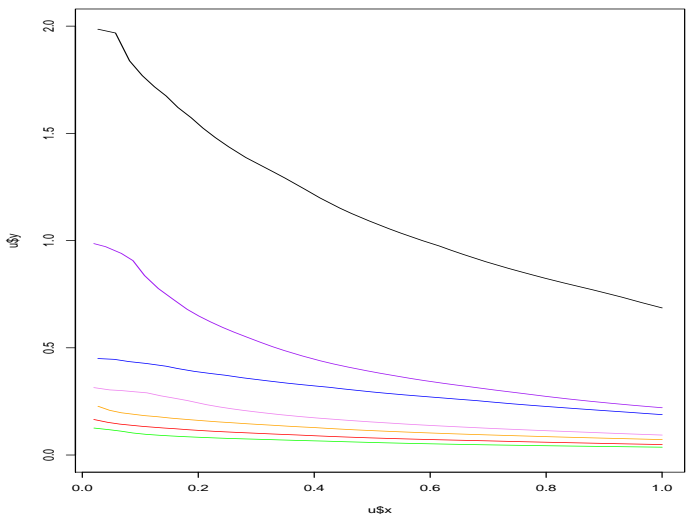}%
\caption{ Lack-of-fit contrast curves on simulated data. Black: constant fit ,
purple: single CART tree, blue: linear model, violet: random forest, orange:
squared-error and red: absolute loss gradient boosting, green: truth.}%
\label{fig20}%
\end{center}
\end{figure}
%EndExpansion

Lack-of-fit contrast curves for six methods are shown in Fig. \ref{fig20}. The
methods are (top to bottom): black constant fit (global mean), purple single
CART tree, blue linear least-squares fit, violet random forest, orange
squared-error and red absolute loss gradient boosting. The bottom green curve
represents the lack-of-fit contrast curve for the true function $f(\mathbf{x}%
)$ on these data. All curves were evaluated on a separate 25000 observation
test data set not used to train the respective models.

\begin{center}
\textbf{Table A1}

RMS estimation error and contrast tree RMS discrepancy for several methods

\qquad

$%
\begin{array}
[c]{ccc}%
\text{Method} & \text{estimation error} & \text{ discrepancy}\\
\text{constant} & 0.99 & 0.86\\
\text{CART tree} & 0.57 & 0.34\\
\text{linear model} & 0.33 & 0.23\\
\text{random forest} & 0.21 & 0.13\\
\text{sqr-error boost} & 0.15 & 0.090\\
\text{abs-error boost} & 0.11 & 0.063\\
\text{truth} & 0 & 0.046
\end{array}
$
\end{center}

Since the data are simulated and truth $f(\mathbf{x})$ is here known one can
directly compute root-mean-squared estimation error%
\[
RMSE=\sqrt{mean((f(\mathbf{x})-\hat{f}(\mathbf{x}))^{2})}%
\]
for each method. This is shown in Table A1 (second column) for each method
(first column). The third column shows the root-mean-squared discrepancy over
the same test observations calculated from the respective contrast trees for
each method. The discrepancy associated with an observation is that of the
contrast tree region that contains it.

Contrast tree discrepancy as computed on the data and estimation error based
on the (usually unknown) truth are seen here to track each other fairly well.
Contrast tree discrepancy is generally smaller that RMS error but relative
ratios of the two between the methods are similar.

It is important to note that contrast trees are not perfect. As with any
learning method they can sometimes fail to capture sufficiently complex
dependencies on the predictor variables $\mathbf{x}$. In such situations
lack-of-fit may be under estimated. Thus contrast trees can reject fit quality
but never absolutely insure it.

\section{ Quantile regression example\label{s43}}

Use of contrast trees in quantile regression is illustrated on the online news
popularity data set described in Section \ref{s44}. Here we apply contrast
trees to diagnose the accuracy of gradient boosting estimates of the median
and $25$th percentile function of $y\,|\,\mathbf{x}$.

The usual quantile regression loss used by gradient boosting for estimating
the $p$th quantile $z$ is given by (\ref{e24.5}) where here $p\in\{0.5,0.25\}$
and $z$ is the corresponding quantile estimate. With contrast trees we use%
\begin{equation}
d_{m}=\left\vert \,p-\frac{1}{N_{m}}\sum_{i\in R_{m}}I(y_{i}<z_{i})\right\vert
\label{e26}%
\end{equation}
as a discrepancy measure. This quantity can be interpreted as prediction error
on the probability scale. It is the absolute difference between the target
probability $p$ and the empirical probability $\Pr(y<z)$ as averaged over the region.

\begin{figure}
[ptb]
\begin{center}
\includegraphics[
height=2.2451in,
width=3.7222in
]%
{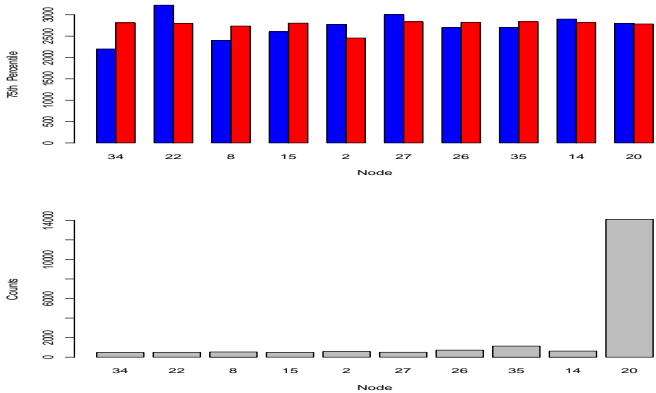}%
\caption{Online news data. Upper frame: Empirical value of the median for
observations in each region (blue), along with the\ corresponding median of
the model predictions (red) in that region, for a quantile contrast tree.
Lower frame: counts in each region.}%
\label{fig3}%
\end{center}
\end{figure}

The data were randomly divided into two parts: a learning data set of $N_{l}=$
$20000$ and and test data set of $N_{t}=19644$. A ten region tree to contrast
the median of $p_{y}(y\,|\,\mathbf{x})$ from its gradient boosting predictions
was built using (\ref{e26}) on the learning data set and evaluated on the test
data set. The results are shown in Fig. \ref{fig3}.%

%TCIMACRO{\FRAME{ftbpFU}{3.7222in}{2.2451in}{0pt}{\Qcb{Online news data. Upper
%frame: Empirical value of the median for observations in each region (blue),
%along with the\ corresponding median of the model predictions (red) in that
%region, for a quantile contrast tree. Lower frame: counts in each region.}%
%}{\Qlb{fig3}}{newsnew.eps}{\special{ language "Scientific Word";
%type "GRAPHIC";  display "USEDEF";  valid_file "F";  width 3.7222in;
%height 2.2451in;  depth 0pt;  original-width 7.9952in;
%original-height 10.5031in;  cropleft "0";  croptop "1";  cropright "1";
%cropbottom "0";  filename 'newsnew.eps';file-properties "XNPEU";}} }%
%BeginExpansion
%EndExpansion

The upper frame shows the empirical (blue) \ and predicted (red) median in
each of the regions in order of absolute discrepancy (\ref{e26}). The lower
frame gives the number of counts in each corresponding region. One sees that
for 85\% of the data (node 20) gradient boosted model predictions of the
median appear to be quite close. In other regions of $\mathbf{x}$ -space there
are small to moderate differences.%

%TCIMACRO{\FRAME{ftbpFU}{3.7429in}{2.7034in}{0pt}{\Qcb{Online news data.
%Lack-of-fit contrast curves comparing conditional median estimates by constant
%(green), linear (red) quantile regression, gradient boosting (black), and
%contrast boosting (blue).}}{\Qlb{fig3.5}}{newscompgof.eps}%
%{\special{ language "Scientific Word";  type "GRAPHIC";  display "USEDEF";
%valid_file "F";  width 3.7429in;  height 2.7034in;  depth 0pt;
%original-width 7.9952in;  original-height 10.5031in;  cropleft "0";
%croptop "1";  cropright "1";  cropbottom "0";
%filename 'newscompgof.eps';file-properties "XNPEU";}} }%
%BeginExpansion
\begin{figure}
[ptb]
\begin{center}
\includegraphics[
height=2.7034in,
width=3.7429in
]%
{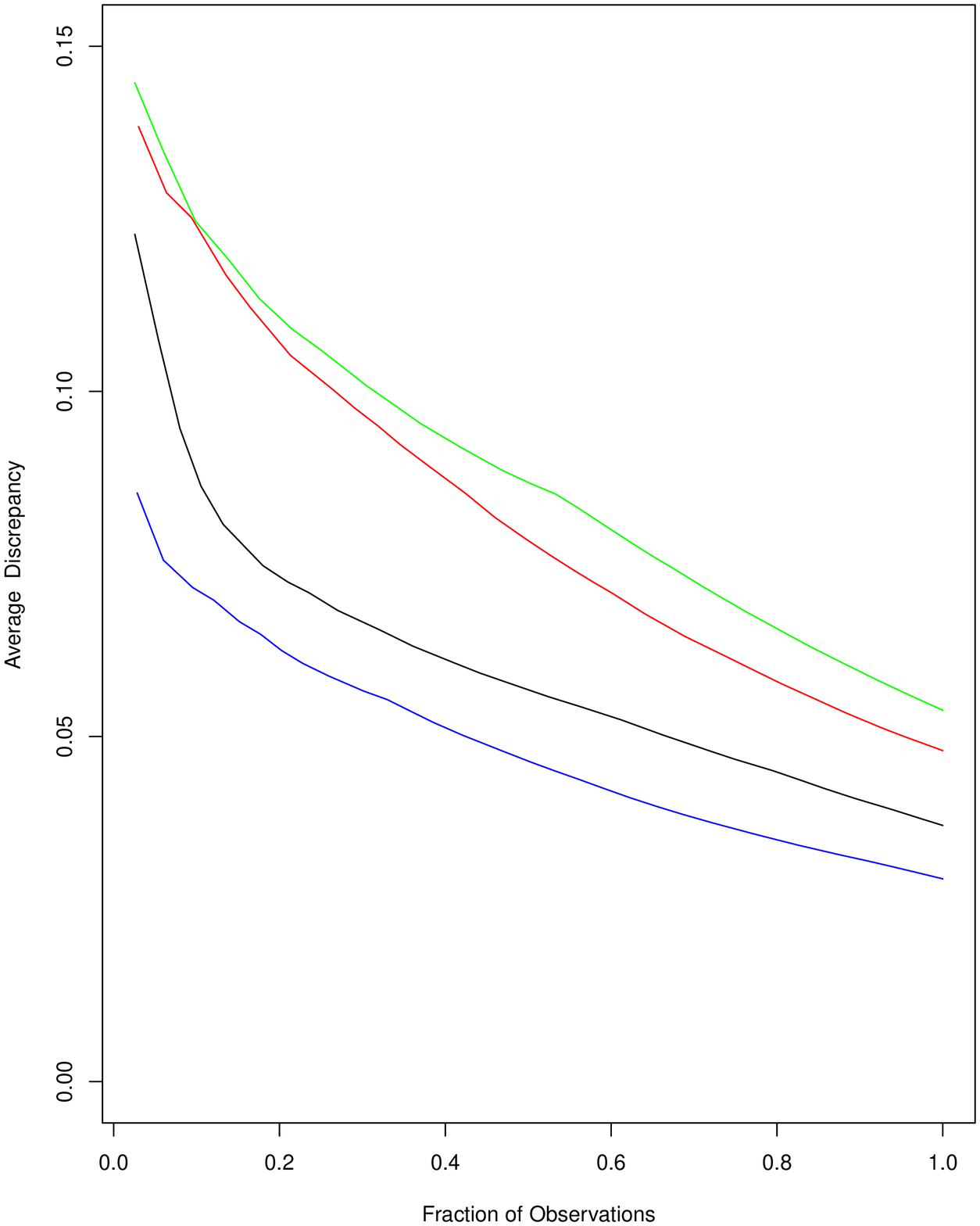}%
\caption{Online news data. Lack-of-fit contrast curves comparing conditional
median estimates by constant (green), linear (red) quantile regression,
gradient boosting (black), and contrast boosting (blue).}%
\label{fig3.5}%
\end{center}
\end{figure}
%EndExpansion

Figure \ref{fig3.5} shows lack-of-fit contrast curves for estimating the
median of $y$ given $\mathbf{x}$ by four methods. The green curve represents a
constant prediction of the global median at each $\mathbf{x}$ - value. The red
curve is for linear quantile regression. The linear model seems only a little
better than the constant one. The black curve represents the gradient boosting
predictions based on (\ref{e24.5}) which are somewhat better. The blue curve
is the result of applying contrast boosting (Section \ref{s5.1}) to the
gradient boosting output. Here this strategy appears to substantially improve
prediction. For the left most points on each curve the bootstraped errors are
respectively $0.015$, $0.016$, $0.018$, and $0.016$ (top to bottom). For the
right most points the corresponding errors are $0.0023$, $0.0026$, $0.0031$
and $0.0033$. Thus, the larger differences between the curves seen in Fig.
\ref{fig3.5} are highly significant.%

%TCIMACRO{\FRAME{ftbpFU}{3.7014in}{2.3488in}{0pt}{\Qcb{Online news data.
%Lack-of-fit contrast curves comparing conditional 25--percentile estimates by
%constant (green), linear (red) quantile regression, gradient boosting (black),
%and contrast boosting (blue).}}{\Qlb{fig3.7}}{newscompgof25.eps}%
%{\special{ language "Scientific Word";  type "GRAPHIC";  display "USEDEF";
%valid_file "F";  width 3.7014in;  height 2.3488in;  depth 0pt;
%original-width 7.9952in;  original-height 10.5031in;  cropleft "0";
%croptop "1";  cropright "1";  cropbottom "0";
%filename 'newscompgof25.eps';file-properties "XNPEU";}} }%
%BeginExpansion
\begin{figure}
[ptb]
\begin{center}
\includegraphics[
height=2.3488in,
width=3.7014in
]%
{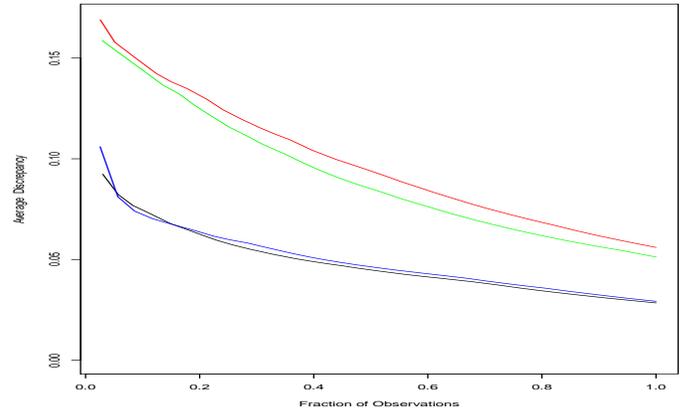}%
\caption{Online news data. Lack-of-fit contrast curves comparing conditional
25--percentile estimates by constant (green), linear (red) quantile
regression, gradient boosting (black), and contrast boosting (blue).}%
\label{fig3.7}%
\end{center}
\end{figure}
%EndExpansion

Figure \ref{fig3.7} shows lack-of-fit contrast curves for estimating the
conditional first quartile ($p=0.25$) as a function of $\mathbf{x}$ for the
same four methods. Here one sees that the global constant fit appears slightly
better that the linear model, while the gradient boosting quantile regression
estimate is\ about twice as accurate. Contrast boosting seems to provide no
improvement in this case. Standard errors on the left most points of the
respective curves are $0.014$, $0.0092$, $0.020$, and $0.015$ (top to bottom).
For the right most curves the corresponding errors are $0.0027$, $0.0039$,
$0.0029$ and $0.0030$.

\begin{center}
\textbf{Table B1}

Prediction risk corresponding to the several quantile regression methods for
online news data

\qquad%

\begin{tabular}
[c]{lll}%
Method & Median & 1st Quartile\\
Constant & 2489.5 & 678.3\\
Linear & 2488.5 & 678.4\\
Gradient Boosting & 2481.7 & 674.1\\
Contrast Boosting & 2479.9 & 674.1
\end{tabular}

\end{center}

Table B1 shows quantile regression prediction risk based on $L_{1}$ loss
(\ref{e24.5}) for median ($p=0.5$) and first quartile ($p=0.25$) using the
four methods shown in Figs. \ref{fig3.5} and \ref{fig3.7}. Although here
prediction risk measures lack-of-accuracy of the methods in the same order as
their respective contrast trees, it gives no indication of their actual
relative or absolute lack-of-fit to the data as seen from their respective
contrast curves in Figs \ref{fig3.5} and \ref{fig3.7}.

\section{Distribution boosting example\label{onpd}}

Distribution boosting is illustrated using the online news popularity data
described in Section \ref{s44}. The goal is to estimate the distribution
$p_{y}(y\,|\,\mathbf{x})$ of ($\log_{10}$) popularity of news articles $y$ for
given sets of predictor variable values $\mathbf{x}$. Here we investigate the
variation of the final distribution estimate $\hat{p}_{y}(y\,|\,\mathbf{x}%
)$\ to different initial $z$ - distributions $p_{z}(z\,|\,\mathbf{x})$. For
the same $p_{y}(y\,|\,\mathbf{x})$, changing the initial $p_{z}%
(z\,|\,\mathbf{x})$ distribution can substantially change the nature of the
target transformation functions $g_{\mathbf{x}}(z)$ to be estimated. This can
affect ultimate accuracy of the estimates $\hat{p}_{y}(y\,|\,\mathbf{x})$.

\begin{figure}
[ptb]
\begin{center}
\includegraphics[
height=2.7527in,
width=3.7152in
]%
{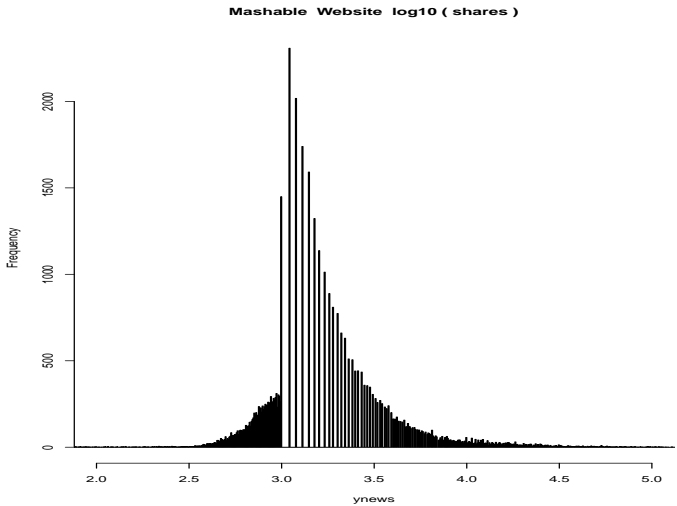}%
\caption{Distribution of $\log_{10}(shares)$ for the online news data.}%
\label{fig16}%
\end{center}
\end{figure}

Distribution boosting was applied to the $20000$ observation training data set
using three different initial $p_{z}(z\,|\,\mathbf{x})$. The first was the
same normal distribution $z\sim N(\bar{y},\sigma_{y}^{2})$ at every
$\mathbf{x}$, where $\bar{y}$ and $\sigma_{y}^{2}$ are the mean and variance
of $y=\log_{10}($popularity$)$. The second initial distribution, also
independent of $\mathbf{x}$, is the empirical marginal distribution of $y$ as
shown in Fig. \ref{fig16}. The third initial $z$ - distribution is that of the
residual bootstrap at each $\mathbf{x}$ as described in Section \ref{s44}.
This assumes homoscedasticity on the log-scale with varying location.%

%TCIMACRO{\FRAME{ftbpFU}{3.7152in}{2.7527in}{0pt}{\Qcb{Distribution of
%$\log_{10}(shares)$ for the online news data.}}{\Qlb{fig16}}{fig6.eps}%
%{\special{ language "Scientific Word";  type "GRAPHIC";  display "USEDEF";
%valid_file "F";  width 3.7152in;  height 2.7527in;  depth 0pt;
%original-width 7.9952in;  original-height 10.2532in;  cropleft "0";
%croptop "1";  cropright "1";  cropbottom "0";
%filename 'fig6.eps';file-properties "XNPEU";}} }%
%BeginExpansion
%EndExpansion

\begin{figure}
[ptb]
\begin{center}
\includegraphics[
height=2.9378in,
width=3.7291in
]%
{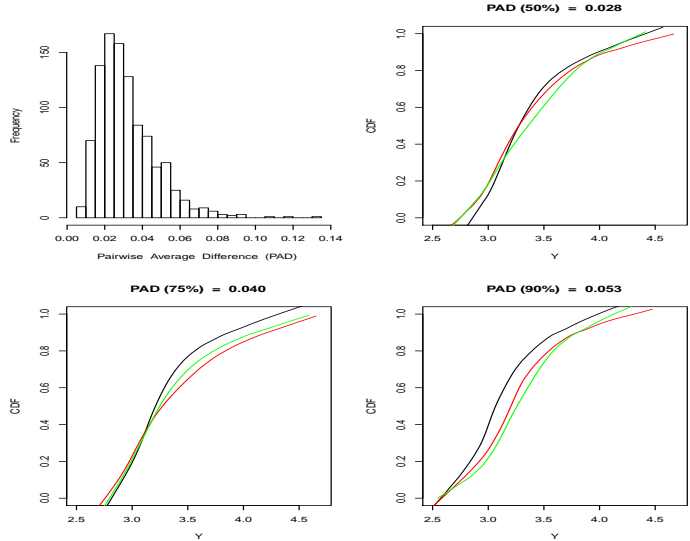}%
\caption{Upper left: distribution of average pair-wise difference between
$CDF$ estimates resulting from the three different initial $z$ - distributions
for online news data. Upper right: CDF estimates for parametric bootstrap
(black), Gaussian (green) and empirical marginal (red) starting distributions
for observation with median pairwise difference. Lower: corresponding plots
for 75\% and 90\% decile difference.}%
\label{fig17}%
\end{center}
\end{figure}

The upper left frame of Fig. \ref{fig17} shows the distribution of the average
pair-wise difference between the three $CDF$ estimates on each (test set)
observation $\mathbf{x}$, resulting from the three different beginning $z$ -
distributions. Difference between two $CDF$ estimates is given by (\ref{e22})
with the $100$ evaluation points $\{u_{j}\}_{1}^{100}$ being a uniform grid
between the minimum of $0.001$ quantiles and the maximum of the $0.999$
quantiles of the three distributions.%

%TCIMACRO{\FRAME{ftbpFU}{3.7291in}{2.9378in}{0pt}{\Qcb{Upper left: distribution
%of average pair-wise difference between $CDF$ estimates resulting from the
%three different initial $z$ - distributions for online news data. Upper right:
%CDF estimates for parametric bootstrap (black), Gaussian (green) and empirical
%marginal (red) starting distributions for observation with median pairwise
%difference. Lower: corresponding plots for 75\% and 90\% decile difference.}%
%}{\Qlb{fig17}}{newsdistcomp.eps}{\special{ language "Scientific Word";
%type "GRAPHIC";  display "USEDEF";  valid_file "F";  width 3.7291in;
%height 2.9378in;  depth 0pt;  original-width 7.9952in;
%original-height 10.5031in;  cropleft "0";  croptop "1";  cropright "1";
%cropbottom "0";  filename 'newsdistcomp.eps';file-properties "XNPEU";}} }%
%BeginExpansion
%EndExpansion

The\ 50, 75, and 90 percentiles of the distribution shown in the upper left
frame are respectively $0.028$, $0.040$, and $0.053$. As in Fig. \ref{fig13}
the remaining plots in Fig. \ref{fig17} show the three corresponding $CDF$s
for those observations with pair-wise average difference closest to these
respective percentiles. The green curves display the estimate corresponding to
the Gaussian starting $z$ - distribution, red for the empirical marginal
distribution of Fig. \ref{fig16}, and black for the residual bootstrap start.
The upper right frame shows that for at least half of the observations the
three estimates are fairy similar. The other half exhibit moderate
differences. The residual bootstrap estimates tend to be different from the
other two, which are usually similar to each other.

Figure \ref{fig17} shows that different starting $z$ - distributions give rise
to at least slightly different conditional distribution estimates. In general,
different methods produce different answers and one would like to ascertain
their respective accuracies. Contrast trees provide a lack-of-fit measure.
With conditional distribution estimates one can contrast $y$ with $\hat
{y}=\hat{g}_{\mathbf{x}}(z)$ on an independent test data set not involved in
the estimation as was illustrated in Fig. \ref{fig5}. Here we employ this
strategy to evaluate the respective accuracies of the three conditional
distribution estimates obtained by the three different starting $z$ - distributions.%

%TCIMACRO{\FRAME{ftbpFU}{3.7438in}{2.9473in}{0pt}{\Qcb{QQ--plots of $y$ versus
%$\hat{y}=\hat{g}_{\mathbf{x}}(z)$ calculated from parametric bootstrap start
%for the nine highest discrepancy regions of a 50 node contrast tree on the
%online news test data set. The red lines represent equality.}}{\Qlb{fig18}%
%}{newsbootqq.eps}{\special{ language "Scientific Word";  type "GRAPHIC";
%display "USEDEF";  valid_file "F";  width 3.7438in;  height 2.9473in;
%depth 0pt;  original-width 7.9952in;  original-height 10.5031in;
%cropleft "0";  croptop "1";  cropright "1";  cropbottom "0";
%filename 'newsbootqq.eps';file-properties "XNPEU";}} }%
%BeginExpansion
\begin{figure}
[ptb]
\begin{center}
\includegraphics[
height=2.9473in,
width=3.7438in
]%
{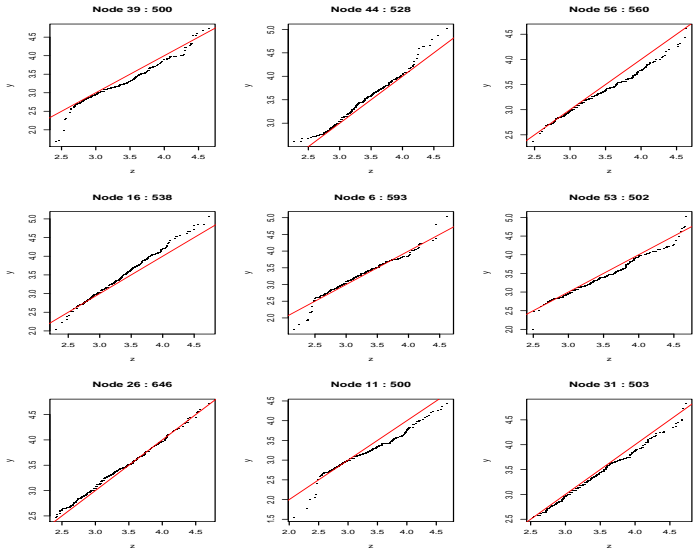}%
\caption{QQ--plots of $y$ versus $\hat{y}=\hat{g}_{\mathbf{x}}(z)$ calculated
from parametric bootstrap start for the nine highest discrepancy regions of a
50 node contrast tree on the online news test data set. The red lines
represent equality.}%
\label{fig18}%
\end{center}
\end{figure}
%EndExpansion

Figure \ref{fig18} shows QQ -- plots of $y$ versus the estimates $\hat{y}%
=\hat{g}_{\mathbf{x}}(z)$ based on the residual bootstrap starting $z$ -
distribution. Shown are the nine largest discrepancy regions of \ a 50
terminal node contrast tree. These nine regions account for 25\% of the data.
This can be compared to Fig. \ref{fig5} which shows the corresponding QQ
--plots for $y$ versus the original residual bootstrap $z$ before distribution boosting.%

%TCIMACRO{\FRAME{ftbpFU}{3.7533in}{2.9637in}{0pt}{\Qcb{ Lack-of-fit contrast
%curves for three trees contrasting $y$ with $\hat{y}=\hat{g}_{\mathbf{x}}(z)$
%based on the different starting $z$ - distributions: Gaussian (green),
%empirical marginal (red) and parametric bootstrap (black). }}{\Qlb{fig18.5}%
%}{newsgof.eps}{\special{ language "Scientific Word";  type "GRAPHIC";
%display "USEDEF";  valid_file "F";  width 3.7533in;  height 2.9637in;
%depth 0pt;  original-width 7.9952in;  original-height 10.5031in;
%cropleft "0";  croptop "1";  cropright "1";  cropbottom "0";
%filename 'newsgof.eps';file-properties "XNPEU";}} }%
%BeginExpansion
\begin{figure}
[ptb]
\begin{center}
\includegraphics[
height=2.9637in,
width=3.7533in
]%
{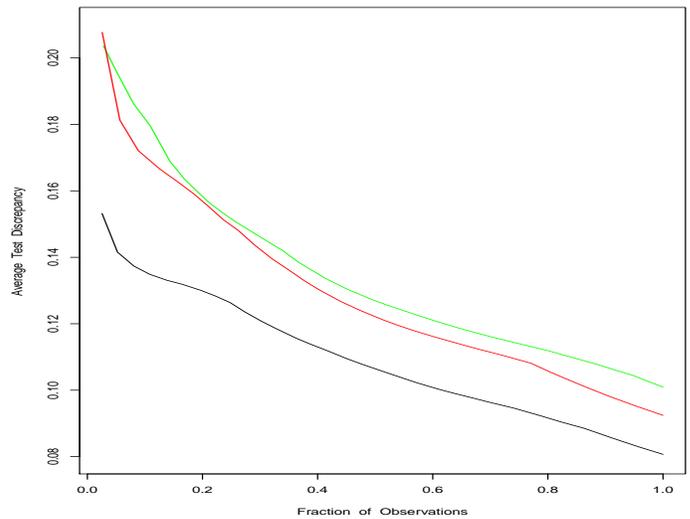}%
\caption{ Lack-of-fit contrast curves for three trees contrasting $y$ with
$\hat{y}=\hat{g}_{\mathbf{x}}(z)$ based on the different starting $z$ -
distributions: Gaussian (green), empirical marginal (red) and parametric
bootstrap (black). }%
\label{fig18.5}%
\end{center}
\end{figure}
%EndExpansion

Figure \ref{fig18.5} shows the lack-of-fit contrast curves corresponding to
the three distribution boosting estimates based on the three different
starting $z$ - distributions. Each line summarizes a different tree
contrasting $y$ with one of the corresponding three estimates $\hat{y}=$
$\hat{g}_{\mathbf{x}}(z)$. The green and red curves in Fig. \ref{fig18.5}
summarize the results for contrasting $y$ with $\hat{g}_{\mathbf{x}}(z)$ based
on the respective Gaussian and empirical marginal distribution (Fig.
\ref{fig16}) starting $z$ - distributions. Their accuracies are seen to be
similar. The black curve summarizes the tree depicted in Fig. \ref{fig18}
contrasting $y$ with the estimates $\hat{g}_{\mathbf{x}}(z)$ based on the
residual bootstrap starting $z$ - distribution. These $\hat{g}_{\mathbf{x}%
}(z)$ estimates appear to be somewhat more accurate. Bootstrap standard errors
on the left most points of all three curves are $0.022$. For the right most
points the corresponding errors are $0.0055$, $0.0052$ and $0.0049$.

The average discrepancy of the tree contrasting $y$ and the residual bootstrap
estimated $\hat{g}_{\mathbf{x}}(z)$ (black) is $0.081$. The corresponding
averages for the respective Gaussian and empirical marginal distribution (Fig.
\ref{fig16}) starting $z$ - distributions are $0.10$ and $0.092$ respectively.
These results can be compared with the discrepancies of their initial
\emph{untransformed} $z$ - distributions. Average discrepancy for contrasting
$y$ with the untransformed residual bootstrap distribution (Fig. \ref{fig5})
is $0.13$. The corresponding average discrepancies with $y$ for the
untransformed Gaussian $z$ distribution is $0.26$, and that for the empirical
marginal distribution is $0.24$. Thus the residual bootstrap provided a much
closer starting point for estimating $p_{y}(y\,|\,\mathbf{x})$ ultimately
resulting in somewhat more accurate results.

One can obtain a null distribution for average transformed discrepancy by
repeatedly applying the contrast boosting procedure with $y$ and $z$ randomly
sampled from the same distribution. In this case $p_{y}(y\,|\,\mathbf{x})$ and
$p_{z}(z\,|\,\mathbf{x})$\ are the same and any differences detected by the
distribution boosting procedure, as revealed by a final contrast tree, are
caused by the random nature of the data and not actual differences between the
respective distributions. Fifty replications of contrasting boosting based on
pairs of random samples, each drawn from from the (same) residual bootstrap
distribution, produced and average tree discrepancy of $0.085$ with a standard
deviation of $0.003$. Thus there is no evidence here for a systematic
difference between the distribution of the original $y$ and its estimate
$\hat{y}=\hat{g}_{\mathbf{x}}(z)$ based on the residual bootstrap initial $z$
- distribution.

\section{Two-sample contrast trees\label{tc}}

Contrast trees as so far described are applied to a single data set where each
observation has two outcomes $y$ and $z$, and a single set of predictor
variables $\mathbf{x}$. A similar methodology can be applied to two--sample
problems where there are separate predictor variable measurements for $y$ and
$z$. Specifically the data consists of two samples $\{\mathbf{x}_{i}%
^{(1)},y_{i}\}_{1}^{N_{1}}$ and $\{\mathbf{x}_{i}^{(2)},z_{i}\}_{1}^{N_{2}}$.
The predictor values $\mathbf{x}_{i}^{(1)}$ correspond to outcomes $y_{i}$,
and the values $\mathbf{x}_{i}^{(2)}$ correspond to $z_{i}$. The goal to to
identify regions in $\mathbf{x}$ - space where the two conditional
distributions $p_{y}(y\,|\,\mathbf{x})$ and $p_{z}(z\,|\,\mathbf{x})$, or
selected properties of those distributions, most differ.

Discrepancy measures for each region $R_{m}$ of $\mathbf{x}$ - space can be
defined in analogy with (\ref{e1})%
\begin{equation}
d_{m}=D(\{y_{i}\}_{\mathbf{x}_{i}^{(1)}\in R_{m}},\{z_{i}\}_{\mathbf{x}%
_{i}^{(2)}\in R_{m}})\text{.} \tag{24}\label{e50}%
\end{equation}
Regions and splits are based on the pooled predictor variable sample
$\{\mathbf{x}_{i}\}_{i=1}^{N}=\{$ $\mathbf{x}_{i}^{(1)}\}_{i=1}^{N_{1}}%
\cup\{\mathbf{x}_{i}^{(2)}\}_{i=1}^{N2}$ with $N=N_{1}+N_{2}$. Tree
construction strategy is the same as that described in Section \ref{s2} using
(\ref{e50}).

We illustrate two--sample contrast trees using the census income data set
described in Section \ref{s41}. One of the samples is taken to be the data
from the 32650 males, and the other sample data from the 16192 females. The
goal is to investigate gender differences in probability of high salary
(greater than \$50K/year, \$100K 2020 equivalent) in terms of the other
demographic and financial variables as reflected in this data set.

The high salary probability averaged over all males in the data set is $0.30$
whereas that for females is $0.11$. Thus the relative odds of high salary for
men is almost three times that for women over the entire data set. Here we use
two--sample contrast trees to investigate whether there are special
demographic and financial characteristics for which these relative odds are
different. Trees were built on a random half sample of 24421 observations and
the corresponding node statistics computed on the other left out half.

We first use two--sample contrast trees to seek regions in predictor variable
$\mathbf{x}$ - space for which male/female relative high salary probability is
larger than 3/1. For this we use a \emph{ratio} discrepancy measure%
\begin{equation}
d_{m}=mean(y_{i}\,|\,\mathbf{x}_{i}^{(1)}\in R_{m})/mean(z_{i}\,|\,\mathbf{x}%
_{i}^{(2)}\in R_{m}) \tag{25}\label{e51}%
\end{equation}
where $\{y_{i},\mathbf{x}_{i}^{(1)}\}_{i=1}^{N_{m}}$ represents the
$N_{m}=32650$ males and $\{z_{i},\mathbf{x}_{i}^{(2)}\}_{i=1}^{N_{f}}$ the
$N_{f}=\ 16192$ females. Here $y_{i}$ and $z_{i}$ are indicators of high (male
and female) salary and $\mathbf{x}_{i}^{(1)},\mathbf{x}_{i}^{(2)}$ are the
corresponding predictor variables.%

%TCIMACRO{\FRAME{ftbpFU}{3.3434in}{3.0856in}{0pt}{\Qcb{Upper frame: probability
%of income greater that \$50K for women (blue) and men (red) in regions
%designed for relatively large values of the latter. Lower frame: Fraction of
%women (blue) and men (red) in each \ region.}}{\Qlb{malefem}}{malefem.eps}%
%{\special{ language "Scientific Word";  type "GRAPHIC";  display "USEDEF";
%valid_file "F";  width 3.3434in;  height 3.0856in;  depth 0pt;
%original-width 7.9952in;  original-height 10.5031in;  cropleft "0";
%croptop "1";  cropright "1";  cropbottom "0";
%filename 'malefem.eps';file-properties "XNPEU";}} }%
%BeginExpansion
\begin{figure}
[ptb]
\begin{center}
\includegraphics[
height=3.0856in,
width=3.3434in
]%
{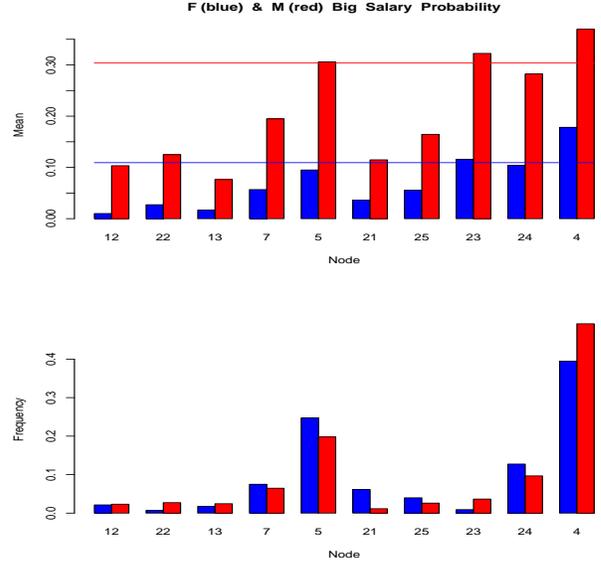}%
\caption{Upper frame: probability of income greater that \$50K for women
(blue) and men (red) in regions designed for relatively large values of the
latter. Lower frame: Fraction of women (blue) and men (red) in each \ region.}%
\label{malefem}%
\end{center}
\end{figure}
%EndExpansion

\ \ Figure \ref{malefem} summarizes results for a ten region contrast tree
using (\ref{e51}). In the top frame the height of blue/red bars represent the
probability of income greater that \$50K for the women/men in the region. In
the bottom frame the blue bars represent the fraction of the 16192 women in
the region whereas red signifies the corresponding fraction of the 32650 men.
The blue/red horizontal lines represent the female/male global average high
salary probabilities.

This contrast tree has found several small regions for which the male/female
odds ratio (\ref{e51}) is much greater than its global average $3/1$. For
example region 12 containing 551 observations has a 10.3/1 ratio. Region 22
with 501 observations has a 4.6/1 ratio. However, in all of the highest ratio
regions the actual male/female probabilities of high salary are well below
their respective global averages. In the higher probability regions the ratios
roughly correspond to the corresponding global averages.

We next attempt to uncover regions in $\mathbf{x}$ - space where the
female/male high salary odds ratio is much greater than its global average of
$1/3$ by using the inverse discrepancy measure%
\[
d_{m}=mean(z_{i}\,|\,\mathbf{x}_{i}^{(2)}\in R_{m})/mean(y_{i}\,|\,\mathbf{x}%
_{i}^{(1)}\in R_{m})\text{.}%
\]
%

%TCIMACRO{\FRAME{ftbpFU}{3.3434in}{3.0303in}{0pt}{\Qcb{Upper frame: probability
%of income greater that \$50K for women (blue) and men (red) in regions
%designed for relatively large values of the former. Lower frame: Fraction of
%women (blue) and men (red) in each \ region.}}{\Qlb{femmale}}{femmale.eps}%
%{\special{ language "Scientific Word";  type "GRAPHIC";  display "USEDEF";
%valid_file "F";  width 3.3434in;  height 3.0303in;  depth 0pt;
%original-width 7.9952in;  original-height 10.5031in;  cropleft "0";
%croptop "1";  cropright "1";  cropbottom "0";
%filename 'femmale.eps';file-properties "XNPEU";}} }%
%BeginExpansion
\begin{figure}
[ptb]
\begin{center}
\includegraphics[
height=3.0303in,
width=3.3434in
]%
{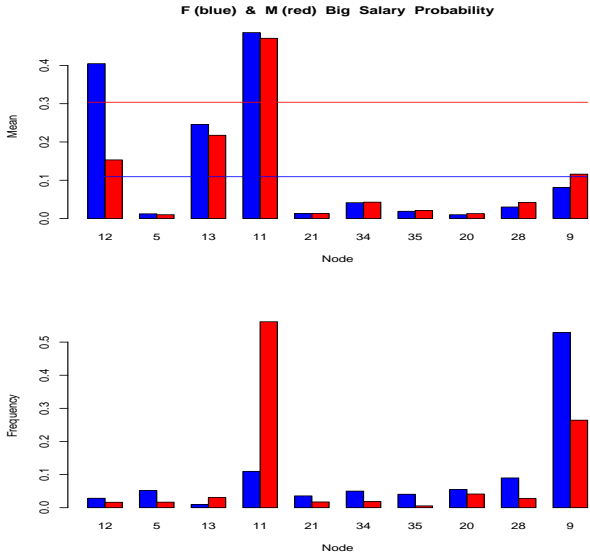}%
\caption{Upper frame: probability of income greater that \$50K for women
(blue) and men (red) in regions designed for relatively large values of the
former. Lower frame: Fraction of women (blue) and men (red) in each \ region.}%
\label{femmale}%
\end{center}
\end{figure}
%EndExpansion
Figure \ref{femmale} summarizes the regions of the corresponding ten region
contrast tree in the same format as Fig. \ref{malefem}. The tree has uncovered
three regions in which the high salary probability for women is higher than
that for men and much higher than its global average (blue line). In region 12
the female/male high salary odds ratio is 2.6/1. In regions 13 and 11 the
probabilities are about equal. In region 11 the overall probability of high
salary for both is relatively very high ($0.47$). This region contains 57\% of
the males and only 11\% of the females in the data set. The rules defining
these three regions are\newpage

\begin{center}
\textbf{Node 12}

\thinspace

$22\leq\,$age\thinspace$<50$

\&

martial status $=$ never married

\&

hours/week $\leq$ $34$

\qquad

\textbf{Node 13}

\thinspace

age $>50$

\&

martial status $=$ never married

\&

hours/week $\leq$ $34$

\qquad

\textbf{Node 11}

age $>22$

\&

martial status $=$ never married

\&

hours/week $>$ $34$
\end{center}

This data set was originally constructed for the purpose of comparing
performance of various machine learning algorithms for predicting high salary.
There is no information as to its representativeness, even for 1994. The
analysis presented here is meant to illustrate the variety of the types of
problems to which contrast trees can be applied.

Contrast trees can be used to compare any two samples based on the same
measured quantities. In particular, the two samples may be taken from the same
system under different conditions or at different times. In these situations
contrast trees can detect the presence of any associated \textquotedblleft
concept drift\textquotedblright\ between the samples, and if detected explain
its nature.

\end{document}